%% file: paper.tex
\documentclass[sigconf]{acmart}
\AtBeginDocument{%
  }

\copyrightyear{2026}
\acmYear{2026}
\setcopyright{cc}
\setcctype{by}
\acmConference[SenSys '26]{ACM/IEEE International Conference on Embedded Artificial Intelligence and Sensing Systems}{May 11--14, 2026}{Saint Malo, France}
\acmBooktitle{ACM/IEEE International Conference on Embedded Artificial Intelligence and Sensing Systems (SenSys '26), May 11--14, 2026, Saint Malo, France}
\acmDOI{10.1145/3774906.3802783}
\acmISBN{979-8-4007-2309-4/2026/05}

\usepackage{microtype}
\usepackage{graphicx}
\usepackage{subcaption}
\usepackage{booktabs} 

\usepackage{hyperref}

\newcommand{\second}[1]{#1}
\newcommand{\add}[1]{#1}

\newcommand\sys{\texttt{PELM}\xspace}

\usepackage{multirow}
\usepackage{xspace}
\usepackage{amsmath}
\usepackage{algorithm}
\usepackage{algorithmic}

\begin{document}

\title[\sys: On-Device LLM Inference with Speculative Decoding and DVFS]{\sys: Power Efficient On-Device LLM Inference with Speculative Decoding and Dynamic Voltage Frequency Scaling}



\author{Weisi Yang}
\affiliation{%
  \institution{Northwestern University}  
  \city{Evanston}
  \state{Illinois}
  \country{USA}
}
\email{weisi@u.northwestern.edu}

\author{Stephen Xia}
\affiliation{%
  \institution{Northwestern University}
  \city{Evanston}
  \state{Illinois}
  \country{USA}
}
\email{stephen.xia@northwestern.edu}


\newcommand{\rev}[1]{#1}
\begin{abstract}
  \input{sections/0-abstract}
\end{abstract}



\maketitle

\input{sections/1-introduction}

\input{sections/2-background}
\input{sections/3-method}

\input{sections/4-evaluation_setup}

\input{sections/5-results}
\input{sections/6-discussion}

\input{sections/8-conclusion}

\bibliographystyle{ACM-Reference-Format}
\bibliography{paper}

\end{document}

%% file: sections/0-abstract.tex
Deploying Large Language Models (LLM) directly on mobile platforms at the edge is gaining a large amount of traction due to a myriad of benefits, such as increased privacy, personalization, and latency. However, LLMs have heavy compute requirements, which are difficult for resource constrained mobile and edge platforms to fulfill. In addition to limited compute resources, mobile and edge systems often have a compact form factor and lack physical mechanisms to dissipate heat generated from high processor usage rates (e.g., fans) to prevent throttling and reduced processing power, which LLMs can easily cause. To mitigate these effects, prior works have proposed various power governing strategies, such as dynamic voltage and frequency scaling (DVFS), for reducing power and heat generation for heavy computational tasks on mobile platforms. Recently, DVFS tailored for mobile LLMs have also been proposed. However, these methods mostly focus on optimizing hardware parameters and processor frequencies, which fall short under some thermally constrained scenarios. Drawing from recent advances in machine learning, we form and take advantage of the key insight that not all tokens require full-depth inference to maintain high quality generation. Motivated by this, we present \sys, a solution that augments traditional DVFS processor frequency tuning with two additional workload-specific knobs: 1) speculative decoding and 2) variable verification depth to expand the optimization space to multiple dimensions for more power efficient on-device LLM inference. In extensive evaluations across different hardware platforms and datasets,reduction in energy consumption, while maintaining comparable task performance. The source code is available at: https://github.com/imec-nu/PELM.

%% file: sections/1-introduction.tex
\section{Introduction}

As mobile devices grow increasingly capable, they are positioned to handle more demanding tasks. The recent development of Large Language Models (LLMs) has attracted significant attention for on-device deployment, driven by the need for ubiquitous computing, data privacy, and offline accessibility \cite{xu2025resource}. Despite increasingly powerful mobile devices, running LLMs locally remains challenging due to constrained hardware resources, resulting in low decoding throughput. Furthermore, the small form factor of these devices limits heat dissipation, making them highly vulnerable to thermal throttling and reduced performance when running high workloads, such as LLM inference, for long periods of time.

Fortunately, there are promising power governing techniques, such as dynamic voltage and frequency scaling (DVFS), which adjust processor voltage and frequency to maintain timely execution of workloads, while preventing excessive power consumption, heat generation, and throttling~\cite{kim2022ztt,lin2023workload}. Tailored DVFS solutions have been designed for conventional deep learning (DL) tasks \cite{geng2024powerlens,zhang2025e4} and more recently for LLMs, optimizing and selecting processor frequencies for different generation patterns \cite{zhang2025dissecting}. However, these hardware-only tuning solutions face a fundamental dilemma: lowering frequency to save power directly throttles throughput and reduces quality of experience (QoE), while raising it exacerbates thermal issues. This makes further optimization of the frequency challenging, especially for LLM workloads where performance and power are highly sensitive to change. To help rectify this problem, our \textbf{key insight} is that running the full model for every token generation is not necessary for maintaining high task performance.

\begin{figure}[t!]
    \centering
    \begin{subfigure}[b]{0.5\columnwidth}
        \centering
        \includegraphics[width=\linewidth]{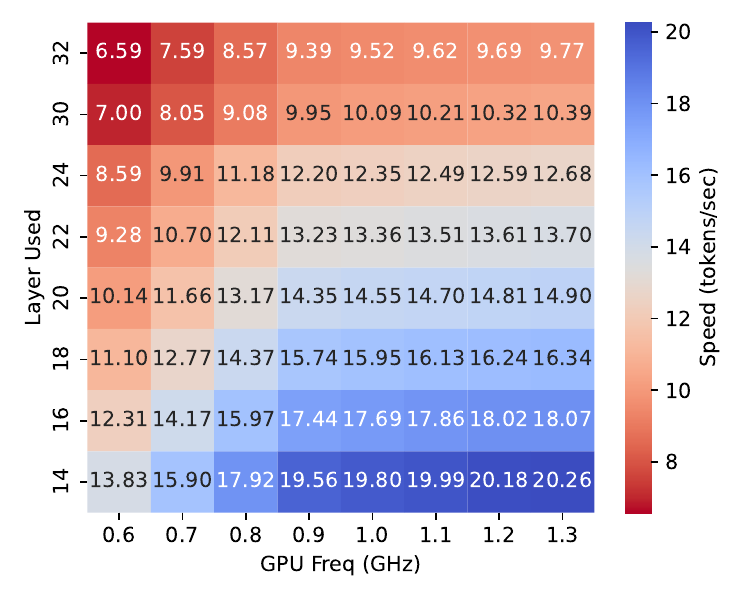}
        \caption{speed}
        \label{fig:speed}
    \end{subfigure}%
    \hfill
    \begin{subfigure}[b]{0.5\columnwidth}
        \centering
        \includegraphics[width=\columnwidth]{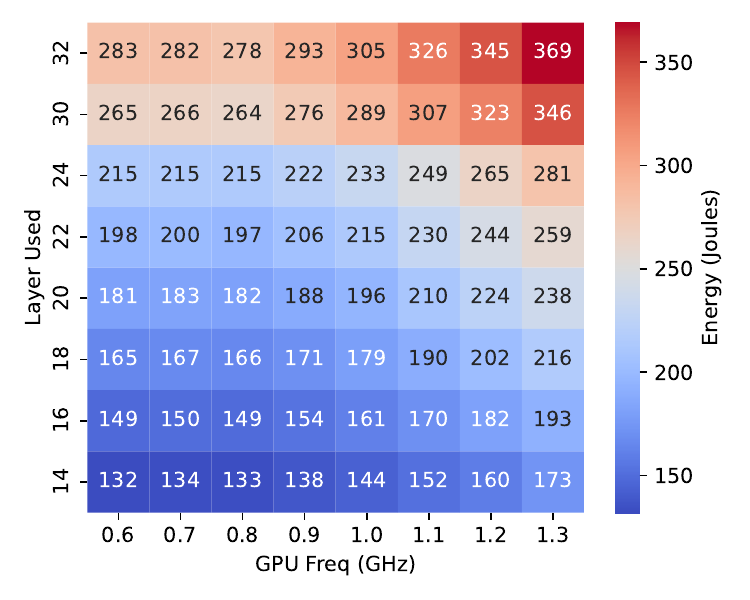}
        \caption{energy}
        \label{fig:energy}
    \end{subfigure}
    \caption{The speed and energy consumption of large language models (LLMs) vary with GPU frequency and the number of layers executed. Tested with LLaMA-3.1-8B on Jetson AGX Orin.}
    \label{fig:pilot-speed-energy-layer-freq}
\end{figure}

\begin{figure}[t!]
    \centering\includegraphics[width=\columnwidth]{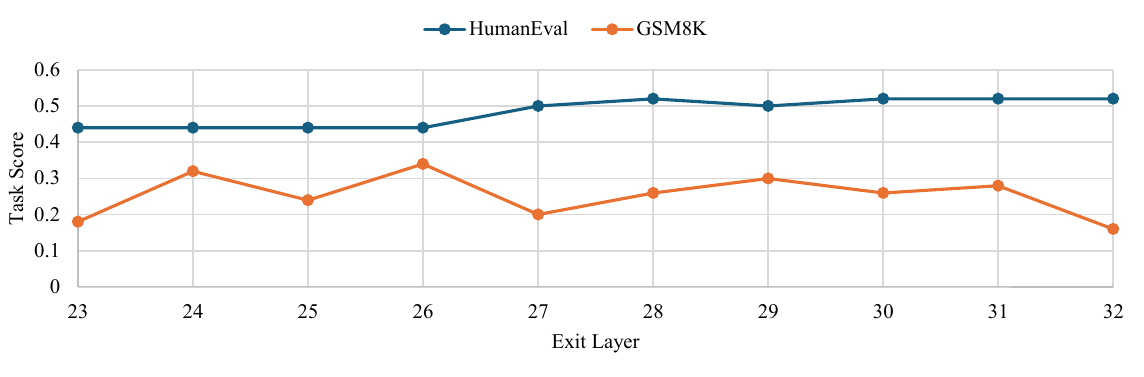}%
\caption{Intermediate layer outputs still convey meaningful information that is comparable to the final output of the model. \rev{Blue: HumanEval (pass@1). Orange: GSM8K (accuracy).} Evaluated on LLaMA-3.1-8B using LayerSkip~\cite{elhoushi2024layerskip}.}\label{fig:pilot-middle-layer}
\end{figure}

For example, speculative decoding leverages this insight for accelerating end-to-end LLM inference without task performance degradation~\cite{zhang2023draft} by using a small, fast ``draft’’ model to generate multiple tokens and then verifying them in parallel with a single pass of the larger and slower ``verify'' model. This allows a single pass of the large model to generate multiple tokens, instead of needing to run separate passes for each token. Self-speculative decoding (SSD) further refines this idea by using a small portion of the model's own intermediate layers for drafting, eliminating the need for a separate model \cite{elhoushi2024layerskip,wei2025adadecode}. However, speculative decoding primarily focuses on improving generation speed, and to the best of our knowledge, has not been explored for power governing on resource-constrained devices.

To see the impacts of varying the number of layers executed and GPU frequency on energy and generation speed, we conducted a pilot study, as shown in Figure~\ref{fig:pilot-speed-energy-layer-freq}, and see that multiple configurations achieve the same QoE (token generation speed). For example, running a 32-layer model at 1.3\,GHz yields a similar speed to a 24-layer model at \rev{0.7}\,GHz, yet the latter consumes 41.7\% less energy. This ``variable’’ depth or ``early-exit’’ execution paradigm has been shown capable of producing similar task performance with much less memory and latency overhead in recent studies~\cite{skean2025layer, elhoushi2024layerskip}. Our own task-specific evaluations in Figure~\ref{fig:pilot-middle-layer} confirm that running a subset of intermediate layers still retains meaningful and usable information for many tokens.

Our key insight exposes a new and larger optimization space by augmenting the traditional control knobs tuned by \textbf{1) DVFS} (hardware frequency) with two additional workload-based knobs: \textbf{2) Self-Speculative Decoding} (algorithmic throughput) and \textbf{3) Variable Depth} (computational workload). However, this creates a \textit{new} and highly complex challenge: a combinatorial control problem. The optimal joint decision for frequency, speculation length, and layer depth is not static; it must ideally adapt \textit{per-token} based on transient thermal conditions and token-specific difficulty \cite{lioubashevski2025looking}. Simple heuristics cannot navigate this dynamic joint state space.

We present \sys, a \underline{P}ower-\underline{E}fficient \underline{L}LM inference framework for \underline{M}obile systems, a power governing solution that jointly optimizes these three components to reduce energy consumption. In addition to tuning hardware frequency parameters, like in traditional DVFS, \sys leverages self-speculative decoding to selectively run a subset of the model’s intermediate layers as the draft model to speed up token generation. Additionally, \sys also introduces an optional third knob (variable depth) that builds on self-speculation, where the number of layers used in the verification model can also be tuned to further reduce energy consumption and latency. First, we formulate this novel joint optimization problem that significantly enlarges the optimization space compared to traditional DVFS methods. Second, to help navigate this large optimization space, we create a Deep Reinforcement Learning (DRL) power governor to tune these knobs and optimize for power and latency at runtime, with comparable task quality as running the full model normally. We conduct comprehensive evaluations with LLMs of different sizes on two Jetson devices under various thermal conditions. These evaluations verify \sys's robustness to different levels of thermal and computational workloads and its adaptability to different QoE requirements, showcasing its feasibility for real-world deployment. The main contributions are as follows:

\begin{enumerate}
    \item We formulate a novel optimization problem for power governing on mobile and edge devices that expands the traditional pathways for limiting heat generation and energy consumption (e.g., tuning processor frequencies in DVFS) during LLM inference, with self-speculative decoding for tuning model throughput and variable depth execution for tuning the computational workload.
    \item We propose \sys, a Deep Reinforcement Learning-based power governing solution that jointly optimizes this formulation at runtime. By incorporating additional algorithmic and workload-based pathways for reducing energy consumption, while maintaining task performance, \sys has more flexibility in how it chooses to limit energy consumption and heat generation than traditional DVFS solutions.
    \item We conduct extensive evaluations on two popular embedded platforms, 5 diverse datasets, demonstrating up to 23.1\% speedup and 52.4\% reduced energy consumption, while maintaining task performance compared to existing state-of-the-art power governors. Our experiments illustrate \sys's effectiveness and robustness across diverse workloads and thermal conditions.

\end{enumerate}

%% file: sections/2-background.tex
\section{Background, Related Works, and Motivation}

\subsection{Dynamic Voltage Frequency Scaling for Edge Devices}

\add{Dynamic Voltage Frequency Scaling (DVFS) is a critical power management technique that adjusts a processor's voltage ($V$) and frequency ($f$) to match the current workload. DVFS leverages the cubic relationship between power and frequency—arising from $P \propto V^2 f$ and the relation $V \propto f$, which yields $P \propto f^3$ \cite{suleiman2005dynamic}—so that small frequency reductions produce significant power savings, making it particularly effective for power-constrained edge devices.
}

 Conventional DVFS implementations (e.g., Linux governors such as schedutil and ondemand) are generic and typically react to coarse-grained metrics like average CPU/GPU utilization, whereas more recent work proposes application-aware DVFS policies that incorporate task-level Quality of Experience (QoE) to jointly optimize hardware and application objectives—for example, maintaining a target FPS in video rendering while minimizing power consumption \cite{kim2022ztt,zhang2024dvfo,yeganeh2020ring,geng2024powerlens,zhang2025e4}. With the rise of on-device LLMs, system-level evaluations have expanded in scope \cite{laskaridis2024melting,li2024large,suleiman2005dynamic,laskaridismobile}, and DVFS schemes tailored to LLM workloads have emerged for both datacenters \cite{kakolyris2024slo,qiu2024power,stojkovic2025tapas,kakolyris2025throttll} and mobile devices \cite{liu2025m,seo2025gold,ye2025agft,liu2025greenllm,zhang2025dissecting}.

\noindent\textbf{Inefficiency of Existing DVFS for LLM Workloads.} \rev{However, both conventional and recent QoE-aware DVFS designs are insufficient for LLM generation due to a fundamental mismatch between their control granularity and the workload characteristics (see Section~\ref{sec-results-dvfs-performance}). Existing DVFS approaches primarily regulate processor frequency, assuming performance scales smoothly with frequency adjustments. In contrast, LLM generation exhibits highly irregular and token-dependent computational patterns, where latency is jointly determined by both hardware frequency and algorithmic factors such as speculative depth and exit strategies. This limited control space of frequency-only adaptation prevents DVFS from fully exploring the performance-power trade-off under stringent QoE constraints. As a result, such methods often make suboptimal decisions in practice: conservative settings may fail to meet latency targets, while aggressive frequency boosting can induce thermal throttling, ultimately degrading sustained performance and increasing energy consumption. These limitations motivate the need for a co-designed optimization strategy tailored to LLM workloads.}

\subsection{Speculative Decoding}

\add{Autoregressive LLM inference is primarily bottlenecked by memory bandwidth, as each token is generated sequentially, leading to underutilization of compute units. Speculative Decoding (SD) \cite{leviathan2023fast} addresses this by using a small "draft model" to generate a $k$-token draft, which is then verified in parallel by the large "target model" in a single, compute-bound pass. This replaces $k$ memory-bound steps with a single step of the target model, significantly accelerating generation.}

The conventional two-model speculative decoding paradigm is impractical on memory-constrained edge devices due to its prohibitive memory overhead. To address this, Self-Speculative Decoding eliminates the auxiliary draft model by reusing a truncated version (e.g., the first $N$ layers) of the target model as a lightweight draft generator \cite{zhang2023draft}. Classic draft–then–verify amortizes a full forward pass over multiple tokens via rejection sampling, achieving 2–4× latency reductions while preserving the original model distribution \cite{leviathan2023fast,zhang2023draft}. However, approaches relying on independent draft backbones incur tokenizer coupling and VRAM overhead, motivating integrated designs such as Medusa (multi-head future token prediction) \cite{cai2024medusa}, \second{EAGLE-1/2/3 (small draft model with multi-layer hidden states fusion, dynamic tree, and training-time test)} \cite{li2024eagle, li2024eagle2, li2025eagle3}, and layer-skipping methods like SWIFT, LayerSkip, and AdaDecode \cite{xia2024swift,elhoushi2024layerskip,wei2025adadecode}, as well as auxiliary-free lookahead schemes \cite{fu2024break}. Building on the LayerSkip-style self-speculative paradigm, we generalize the design by allowing both draft exit depth and verification depth to vary dynamically at each decoding step, while co-optimizing speculative length under a DVFS-aware governor. This joint control sustains QoE while respecting power and thermal constraints on mobile hardware.

\subsection{Observations}

\add{\rev{Here we leverage the key insight:} not all tokens require full depth inference, and those processed by shallower depth still maintain useful information while excelling in terms of latency and power consumption. Such insight primarily draws from recent advances in the machine learning field about the study of intermediate layers of LLMs \cite{schuster2022confident, skean2025layer}, indicating that the output of final layers is not necessarily better than that of the earlier layers.}

\add{To validate this, we profiled the LLaMA-8B on an AGX Orin across a 2D space of GPU frequency and executed layer depth (Figure~\ref{fig:pilot-speed-energy-layer-freq}). We first observe a crucial many-to-one mapping: numerous (Layer, Frequency) combinations achieve the same QoE (token speed), yet have vastly different energy costs. \rev{For example, (20 Layers, 0.6 GHz) matches the ~10 tokens/s speed of (32 Layers, 1.3 GHz) but consumes ~50.9\% less energy. This reveals a massive, untapped optimization space.}}

\add{This optimization opportunity is only viable if executing fewer layers does not catastrophically degrade task performance. And Figure~\ref{fig:pilot-middle-layer} confirms that task performance is surprisingly substantial between intermediate (e.g., layer 24) and final layers, retaining meaningful information. \rev{While GSM8K exhibits larger fluctuations across layers due to its sensitivity to reasoning consistency, overall performance remains within a comparable range to the final layer.} Together, these findings form our core motivation: shallower execution is feasible (quality is preserved) and essential (it saves energy for the same QoE). Existing DVFS methods are blind to this, as they only control the "frequency" knob. This necessitates a new controller that can \textbf{co-manage both layer depth and frequency} to find the better, energy-optimized state for any QoE target.}

\subsection{\rev{Positioning of \sys}}

\rev{\sys bridges two previously separate lines of work: speculative decoding in the ML community and DVFS-based power management in systems research.}

\rev{On one hand, recent SD methods \cite{li2024eagle, xia2024swift, elhoushi2024layerskip} focus on reducing latency by adapting model execution depth or by using lightweight draft generators. For example, the EAGLE series improves speculative decoding by attaching lightweight draft heads to the target model's hidden states and predicting future tokens, with later versions introducing dynamic drafting and multi-layer feature fusion to improve acceptance rates. These approaches primarily target throughput improvement under fixed hardware configurations rather than energy-constrained QoE control. On the other hand, conventional DVFS schemes adjust hardware frequency based on coarse utilization signals, without visibility into model-level execution depth or decoding dynamics. As a result, they treat the LLM workload as a fixed black box and lack fine-grained control over computational intensity.}

\rev{\sys operates at the intersection of these two spaces. It jointly manages hardware knobs (DVFS frequency) and software decoding parameters (self-speculation depth, verification depth, and speculative horizon) under explicit thermal and power constraints. By coordinating workload shaping with hardware adaptation in a closed loop, \sys opens a new cross-layer optimization space that is inaccessible to decoding-only or hardware-only approaches.}

%% file: sections/3-method.tex
\section{Method}

\begin{figure*}[t!]
    \centering\includegraphics[width=2\columnwidth]{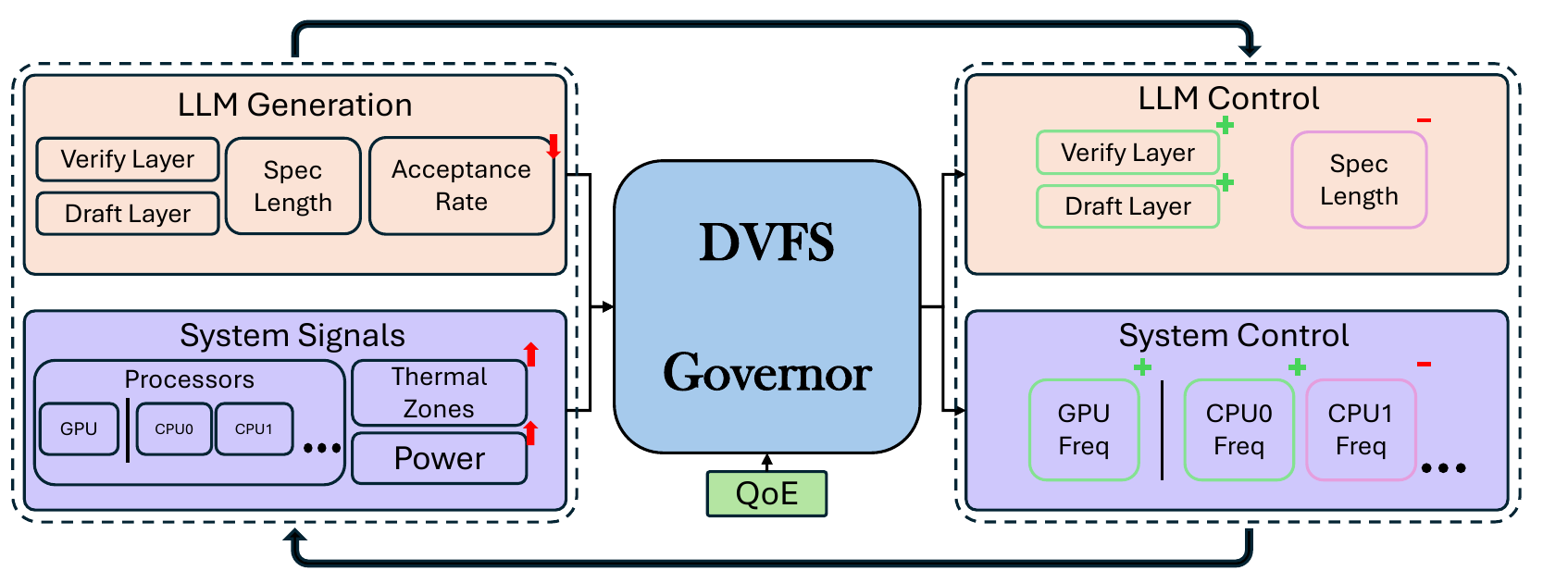}%
\caption{\sys~workflow. At each control step, the governor monitors system and decoding signals (utilization, temperature, power, token/s, and acceptance statistics) and selects joint actions: DVFS settings and decoding knobs (exit layer, speculation length, and verification depth). The next window executes with early-exit drafting and variable-depth verification, supported by HQCache for efficient layer skipping. This closed loop co-optimizes hardware frequency and decoding depth to maintain QoE while reducing energy and avoiding thermal throttling, adapting dynamically across workloads.}\label{fig:workflow}
\end{figure*}

In this section, we present the design of \sys, with its workflow illustrated in Figure~\ref{fig:workflow}. The central component of \sys is the DVFS Governor, which continuously monitors system metrics, including both hardware parameters and software execution metrics, and provides instructions for subsequent LLM generation and hardware configuration. We next formulate the optimization problem that the DVFS Governor is designed to address.

\subsection{Problem Formulation}
\label{sec-method-problem}

The primary consideration for on-device LLM inference is Quality of Experience (QoE) of the user, which we define as the token decoding speed. In addition to QoE, the process should also address efficiency requirements. Our design principle is to provide a power-efficient service that meets the QoE target while minimizing side effects related to power consumption and overheating. Traditional DVFS solves the following problem to optimize these objectives.

\noindent\textbf{Standard DVFS formulation.} Let $q$ be an input query to an LLM, $Q$ be a set of queries, $J$ be energy consumed, $t$ be the time, $T_{temp}(t)$ be the temperature at time $t$, $S$ be the decoding speed, the optimization problem could be:

\begin{equation}
\label{eq:op1}
\begin{aligned}
    \min \quad & E_{q \sim Q}[J(q)] \\
    \text{s.t.} \quad & T_{\text{temp}}(t) < T_{\text{limit}} \\
    & S(q) \ge S_{\text{target}}
\end{aligned}
\end{equation}

The optimization problem (\ref{eq:op1}) aims at minimizing the expected energy consumption throughout the LLM inference process while meeting thermal and QoE constraints. Since heating is fundamentally related to power dissipation, which is the rate of energy consumption, we could rewrite the objective function as $\min E_{q \sim Q}[p(q)\times{t(q)}]$, where $p(q)$ is the power consumption of a query $q$, and $t(q)$ indicates the time used for processing $q$.

\noindent
\textbf{Power and Latency Dilemma.} Minimizing either $p(q)$ or $t(q)$ is viable for optimizing the formulation in Equation~\ref{eq:op1}. However, these two directions indicate a contradiction that makes this optimization challenging to solve. For minimizing $p(q)$, given that a processor's power $p$ is proportional to its executing frequency $f$, and more precisely $p \propto V^2\times{f}$, where $V$ is the working voltage. Since $V \propto f$ \cite{suleiman2005dynamic}, we have $p \propto f^3$. This suggests that $p$ could be minimized by reducing operating frequency $f$. Conversely, to minimize $t(q)$, we maximize generation speed $S(q)$ to reduce the time needed, which requires raising the frequency $f$, contradicting the former optimization. This inherent $p(q)$ and $t(q)$ trade-off must be systematically addressed. We incorporate self-speculative decoding and variable depth execution as two additional knobs for the power governor to minimize execution time, $t(q)$, discussed next.

\noindent\textbf{Self-speculative decoding with variable depth.} Unlike conventional speculative decoding, which requires distinctive draft and verify models, self-speculative decoding uses a sub-part of the full model for drafting, providing a memory-efficient way for speculative computing. Let $ex$ be the depth of draft model, $L$ be the depth of the full model, $ns$ be the number of speculations made. For one time speculation, we execute $ns$ times of $ex$ layers of an LLM to get $ns$ draft tokens, and execute a full-depth run to verify these draft tokens, so the number of total layers run is $ns\times{ex}+L$. Depending on the draft quality, the number of accepted tokens is $n_{ac}$, $n_{ac} \in \{1,2,...,ns+1\}$. Theoretically we could compute the speedup ratio $R$ as $R = {L/((ns\times{ex}+L)}/{n_{ac})} = (n_{ac}\times{L})/(ns\times{ex}+L)$. Then the token decoding speed could be expressed as $S = R\times{S_{Base}}$, where $S_{Base}$ is the speed of running full model depths.

From our initial experiments (Figure~\ref{fig:pilot-middle-layer}), showing that early-exiting at an LLM's intermediate layer often attains comparable output quality as running up to the final layer, we further introduce self-speculative decoding with variable depth. Let $vd$ indicate the variable depth and $ns<{vd}\le{L}$ in a single speculation. Only $vd$ layers of original model act as the verify model instead of full depth, so the number of total layers run becomes $ns\times{ex}+vd$. Now, the speedup ratio is calculated as $R = (n_{ac}\times{vd})/(ns\times{ex}+vd)$.

\noindent\textbf{\sys's optimization formulation.} We can split the objective function $\min E_{q \sim Q}[p(q)\times{t(q)}]$ without introducing contradictions between maximizing and minimizing $f$. We minimize $f$ specifically for $\min E_{q \sim Q}[p(q)]$ and maximize the speedup ratio from speculative decoding $R$ for $\min E_{q \sim Q}[t(q)]$. Moreover, since $vd$ is involved, meaning that the verification model could run partially and produce responses that diverge in target distribution from the full model, we add a new QoE objective -- output fidelity $\phi$, which describes the degree to which the generated content matches the output of the original full model. $vd$, or the depth of the verification model, is a natural proxy for this new metric. This metric provides a new objective for maximizing fidelity $\phi$, which encourages LLM to run full depth. We also model power $p$'s relationship to $f$ as $p(f)$. \rev{We scalarize the multi-objective formulation via a weighted summation, where $\lambda_p$, $\lambda_\phi$, and $\lambda_R$ control the trade-offs among power, fidelity, and speedup.} \second{Additional practical considerations such as thermal limits and utilization–frequency matching are incorporated in the reward design, as described in the next subsection.}

Given new and replaced objectives, the original problem with variable depth self-speculation can be rewritten as: 

\rev{
\begin{equation}
\label{eq:op2}
\begin{aligned}
\min \quad 
& E_{q \sim Q} \Big[
\lambda_p p(f,q)
+ \lambda_\phi (1-\phi(vd|q))
+ \lambda_R \big(-R(ex,ns|vd,q)\big)
\Big] \\
\text{s.t.} \quad 
& T_{\text{temp}}(t) < T_{\text{limit}} \\
& S(q) \ge S_{\text{target}}
\end{aligned}
\end{equation}
}

As such, the original problem (\ref{eq:op1}) is reshaped into a concrete and practical problem (\ref{eq:op2}) with multiple objectives that incorporates both hardware and algorithmic (self-speculative decoding) configurations. Taking inspiration from recent DVFS methods, problem (\ref{eq:op2}) could further be solved in a reinforcement learning manner~\cite{kim2022ztt,lin2023workload,zhang2024dvfo}, we next further formalize our problem within the Q-learning context, a widely used RL algorithm.

\noindent\textbf{Convert to a Q-learning problem.} The problem (\ref{eq:op2}) objective could be replaced by

\begin{equation}
\label{eq:op3}
\begin{aligned}
    \max_{\pi} \quad & E_{\pi}[Q_{\pi}(s,a)] \\
\end{aligned}
\end{equation}

where, $s$ is the state of the environment, and $a$ is the action taken by the agent. $Q_{\pi}(s,a)$ is a value function of state $s$ and action $a$ (details in Section~\ref{sec-method:s-a-r-design}) given a policy $\pi$. The Q-function is iteratively updated via the Bellman Equation: $Q(s_t, a_t) \leftarrow Q(s_t, a_t) + \alpha \left( r_{t+1} + \gamma \max_{a'} Q(s_{t+1}, a') - Q(s_t, a_t) \right)$ to find an optimized policy or Q-function, where $r$ is the reward observed given current state $s_t$ and action $a_t$ taken, $\gamma$ is a discounting factor, and $\alpha$ is the learning rate. As a model-free RL framework, Q-learning is free from establishing state transition models. For problems with small state and actions spaces, the Q-function can be stored in a lookup table (e.g., Q-Table). However, in cases where the state and action spaces are continuous or large, using a Q-Table to store Q-values used for evaluating the Q-function is prohibitively expensive. As such, we approximate the Q-function with a neural network or a Deep Q-Network (DQN). We detail the implementation of our DQN in Section~\ref{sec-method-implementation}. \rev{Here, the Q-function is defined over a reward signal that instantiates 
the scalarized objective in Eq.~(\ref{eq:op2}). 
Specifically, the per-step reward is constructed as the negative 
weighted objective, such that maximizing the expected Q-value 
approximates minimizing Eq.~(\ref{eq:op2}). 
Therefore, the RL formulation is a direct optimization surrogate 
of the original control objective.}

Next, we discuss the state, action, and reward of our DRL problem.

\subsection{State, Action, and Reward Design}
\label{sec-method:s-a-r-design}

The states, actions and rewards used to solve problem (\ref{eq:op3}) are listed in Table~\ref{tab:rl_symbols}, which we detail in the following section.

\begin{table}[]
    \centering
    \caption{RL components and symbols for \sys}
    \resizebox{\columnwidth}{!}{
    \begin{tabular}{l c}
        \toprule
        \textbf{State $s$} &  $p_{c}$, $p_{g}$,$u_{c}$, $u_{g}$, $T_{temp}$,$\delta_{temp}$,$S$, $vd$, $R(ex,ns | vd)$,    $f_{c}$, $f_{g}$ \\
        \midrule
        \textbf{Action $a$} & $f_{c}$, $f_{g}$, $vd$, $ex$, $ns$ \\
        \midrule
        \textbf{Reward Function $r$} & $r_S+r_{vd}+r_{thermal}+r_{u}+r_{p}$\\
        \bottomrule
    \end{tabular}
    }
    \label{tab:rl_symbols}
\end{table}

\subsubsection{State}

The state space of~\sys consists of power consumption of CPU and GPU -- $p_{c}$ and $p_{g}$, processor utilization for CPU and GPU -- $u_{c}$ and $u_{g}$, current temperature and its rate of change -- $T_{temp}$ and $\delta_{temp}$, the self speculation parameters -- variable depth $vd$, the speculation speedup ratio $R(ex,ns | vd)$ as computed in Section~\ref{sec-method-problem}, and current CPU \& GPU frequency -- $f_{c}$ and $f_{g}$.

\subsubsection{Action}

On the hardware side, available actions are CPU \& GPU operating frequency $f_c$ and $f_g$. On the speculation side, we have exit layer $ex$, number of speculations $ns$, and variable depth $vd$. \sys takes $\epsilon$-greedy strategy to decide between exploration and exploitation. It leverages actions generated by DQN with probability $\epsilon$, and does exploration with probability $1-\epsilon$. Next we describe the behavior policy when \sys is in exploration phase.

\noindent\textbf{Behavior policy.} Instead of adopting random sampling from action space for exploration~\cite{kim2022ztt}, we provide a simple yet effective knowledge-guided policy design, directly inspired from our optimization problem (Section~\ref{sec-method-problem}). The behavior policy operates on few key metrics: 1) thermal headroom $t_h$, which is the gap between current temperature $t_{temp}$ and predefined warning temperature $t_{warn}$; 2) speedup ratio: $S/S_{target}$; 3) speculative speedup ratio $R(ex, ns | vd)$.

Using these metrics, the agent performs an informed action. When it observes a sufficient thermal headroom, it takes aggressive actions, such as increasing $f_c$ and $f_g$ to make decoding speed $S$ to match $S_{target}$. When it observes overspeed (i.e. $S>1.2\times{S_{target}}$), it reduces the current frequency to mitigate excessive power consumption and overheating. In scenarios where thermal headroom is limited, the agent adjusts the frequencies more conservatively. Specifically, it takes a smaller step to raise frequency when underspeed is observed and a larger step to lower the frequency for overspeed, and also considers reducing variable depth $vd$ for faster LLM inference, at the cost of output fidelity. This idea of gradually increasing frequency until an ``error'' (e.g., overspeed) occurs, and then backing off at an increase rate, is inspired by congestion control algorithms, such as additive increase multiplicative decrease (AIMD), that have shown stable, fair, and efficient resource allocation behavior~\cite{dumas2002markovian}.

For self-speculation, a feedback loop is employed for exploration. By comparing current speculation speedup ratio $R_t(ex_t, ns_t | vd_t)$ with the previous time step ratio $R_{t-1}(ex_{t-1}, ns_{t-1} | vd_{t-1})$, the agent adjusts its next speculation strategy accordingly. For instance, if the current ratio is higher than previous one, it reduces $ex$ for shorter draft stage time, and increases the number of speculations $ns$ to generate more draft tokens in a single step. Conversely, if the ratio is lower, the agent performs the reverse actions. This knowledge-guided behavior policy enables efficient decision-making to achieve optimized QoE while satisfying efficiency and other constraints.

\subsubsection{Reward}

We describe each term of reward function in Table~\ref{tab:rl_symbols}. To meet decoding speed QoE demand, we define speed ratio $S_r=S/S_{target}$, then speed reward is defined as

$$r_S = 
\begin{cases}
\alpha_{S}(S_r - 0.95) & \text{if } S_r < 0.95  \\
(S_r - 0.95)(1-f_{r-gpu})R & \text{if } 0.95 \le S_r \le 1.2  \\
(1.45 - S_r)(1-f_{r-gpu})R & \text{if } 1.2 < S_r \le 1.45  \\
(1.45 - S_r)(f_{r-gpu})R & \text{if } S_r > 1.45 
\end{cases}$$

where $f_{r-gpu}$ is the current GPU frequency ratio, $R$ is speculation speedup, $\alpha_S$ is a penalty factor. The speed within the range of 0.95x to 1.2x the target speed are considered acceptable, allowing for minor underspeed and moderate speedup. Both overspeed and underspeed have decreasing or even negative reward value. It also encourages to use as low GPU frequency as possible and rewards high speculation speedup. By doing this, the agent manages to maintain speed at target level with relatively low GPU frequency. $r_{vd}$ refers to variable depth reward, defined as:

$$r_{vd} = \begin{cases}
\alpha_{vd+} \cdot F_t \cdot (1 - vd_r) & \text{if } F_t > 0 \\
\alpha_{vd-} \cdot F_t \cdot (1 - vd_r) & \text{if } F_t \le 0
\end{cases}$$

where $vd_r$ is ratio between variable depth and full depth of an LLM, $\alpha_{vd+}$ and $\alpha_{vd-}$ are reward factors, $F_t = \text{tanh}(T_{temp} - T_{vd})$, $T_{vd}$ is a predefined threshold temperature. This term rewards deep depth $vd$ when thermal headroom is adequate, while promoting the use of shallower models when thermal headroom is limited.

For thermal consideration, \sys~employs a sigmoid function that gives reward based on the gap between current temperature and warning temperature, plus a rate of change consideration: $r_{thermal} = - (\text{sigmoid}(T_{temp}-T_{warning})+\delta_{temp})$. When the temperature is significantly below the warning threshold, the penalty approaches zero; however, as $T_{temp}$ approaches and exceeds the warning threshold, the penalty increases  progressively.

\sys~also penalizes mismatched workloads and processor frequencies, denoted as $r_u$. For each processor $proc_i$, $i \in \{1,2,..,N\}$, the corresponding penalty value is:

$$p_{ui} = \begin{cases}
u - f_{ri} & \text{if } u > f_{ri} + 0.05 \\
0 & \text{otherwise}
\end{cases}$$

where $f_{ri}$ is the frequency ratio for $proc_i$. Then $r_u = -\sum_{i=1}^{N}{p_{ui}}$, which is the sum of each processor's penalty. Lastly, we have term $r_p$ that penalizes large power consumption: $r_p = - \alpha_{p} \times \sum_{i=1}^{N}(p_i)$.

\begin{algorithm}[t]
\caption{PELM Runtime Control}
\label{alg:pelm}
\begin{algorithmic}[1]

\STATE Initialize DQN parameters $\theta$, target network $\theta^{-}$, and replay buffer $\mathcal{D}$
\STATE Observe initial system state $s_1$
\STATE Initialize DVFS settings and decoding configuration

\FOR{$t = 1,2,\ldots$}

    \STATE Select action $a_t \sim \pi_\theta(s_t)$

    \STATE Update CPU/GPU frequencies $(f_{c0}, f_{c1}, f_{c2}, f_g)$
    \STATE Set decoding parameters $(e_t, n_t, v_t)$

    \STATE Execute system for control interval $\Delta t$

    \STATE Observe next state $s_{t+1}$

    \STATE Compute reward $r_t = \mathcal{R}(s_t, a_t, s_{t+1})$

    \IF{system is in valid operating regime}
        \STATE Store $(s_t, a_t, r_t, s_{t+1})$ in $\mathcal{D}$
    \ENDIF

    \STATE Update $\theta$ using minibatch from $\mathcal{D}$

    \STATE Update target network $\theta^{-} \leftarrow \theta$ (periodically)

    \STATE $s_t \leftarrow s_{t+1}$

\ENDFOR

\end{algorithmic}
\end{algorithm}

\add{The complete updating and control procedure of \sys is presented in Algorithm~\ref{alg:pelm}.}

\subsection{Saving Compute Caused by Missing KV-Cache}

\begin{figure}[t!]
    \centering\includegraphics[width=\columnwidth]{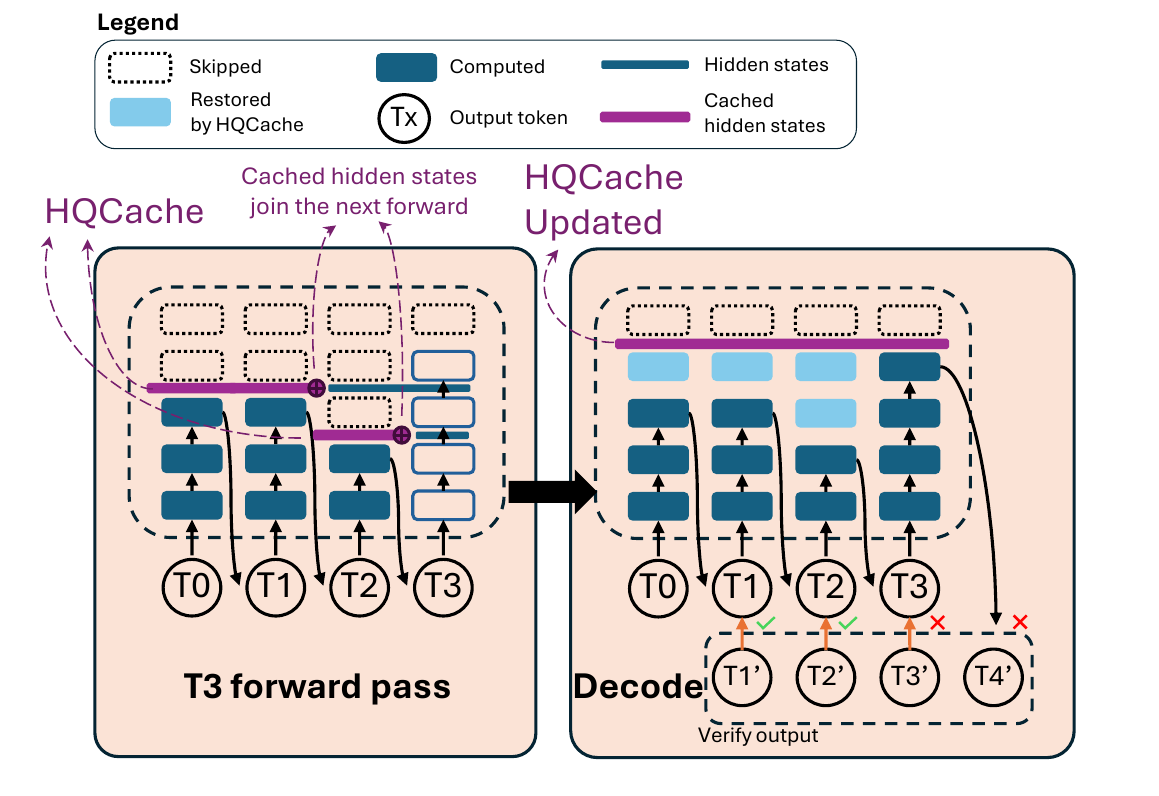}%
\caption{Hidden Queue Cache mechanism to minimize overhead for recomputing KV Cache at deeper layers for previously generated tokens.}\label{fig:hqcache}
\end{figure}

Due to variable depth, each time the LLM may use different depths instead of conventional full model for verification. This causes an issue illustrated in Figure~\ref{fig:hqcache}, when the 4th token has been decoded and its target depth exceeds previous tokens, KV-cache missing happens in layers where previous token decodings did not happen. A naive approach is to concatenate the previous token to recompute from the first layer to the last token's target layer, which is computationally inefficient. Drawing ideas from previous work \cite{elhoushi2024layerskip,wei2025adadecode}, we propose to use Hidden Queue Cache (HQCache) for storing hidden states of intermediate layers. When the target depth for generating the current token exceeds the depth used to generate the previous token, we can load the hidden state and KV-cache from the deepest layer used to generate the previous tokens and recompute from there rather than from the first layer. Prior work leverages this idea to cache and reload from select layers to speed up early exit strategies. In this work, we generalize this idea to cache and reload from any layer and apply to both the drafting and verification steps in self-speculative decoding.

\subsection{Implementation}
\label{sec-method-implementation}

\begin{table}[t]
\centering
\caption{Hardware platforms description.}
\resizebox{\columnwidth}{!}{
\begin{tabular}{lcc}
\toprule
\textbf{Hardware} & \textbf{Jetson AGX Orin} & \textbf{Jetson Orin Nano} \\
\midrule
CPU & 12$\times$ Arm Cortex-A78AE & 6$\times$ Arm Cortex-A78AE \\
CPU Frequency & 0.1--2.0 GHz (29 levels) & 0.1--1.7 GHz (22 levels) \\
CPU Clusters & 3 clusters & 2 clusters \\
GPU & Ampere (2048 CUDA cores) & Ampere (1042 CUDA cores) \\
GPU Frequency & 0.3--1.3 GHz (11 levels) & 0.3--1.0 GHz (8 levels) \\
Memory & 64 GB LPDDR5 & 8 GB LPDDR5 \\
Typical Power & 15--60 W & 7--15 W \\
\bottomrule
\end{tabular}
}
\label{tab:hardware}
\end{table}

\noindent\textbf{Hardware platforms.} We deploy \sys~on two widely used embedded edge devices -- Jetson AGX Orin and Jetson Orin Nano, representing high-end and lower-end mobile computing solutions respectively. The detailed hardware specifications are summarized in Table~\ref{tab:hardware}. Both devices are installed with Jetpack 6.2.1, running on Ubuntu 22.04.5 LTS. We use Python 3.10.12 for the DVFS monitoring and controlling, leveraging \textit{sysfs}~\cite{mochel2005sysfs} for system metrics and frequency control.

\begin{figure}
    \centering\includegraphics[width=0.9\columnwidth]{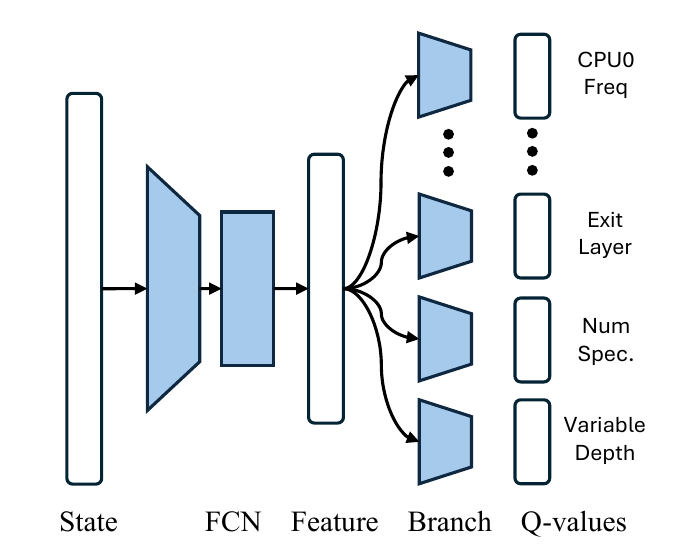}%
\caption{Branch Q-learning network structure.}\label{fig:bdqn}
\end{figure}

\noindent\textbf{Q-learning implementation.} Given the large action space—such as up to 29 frequency configurations for each CPU cluster and 11 for the GPU on the NVIDIA Jetson AGX Orin—we adopt a branch-DQN framework to address the complexity, following the methodology proposed in prior work~\cite{lin2023workload}. Specifically, the high-dimensional action space is decomposed into multiple sub-domains, significantly reducing both model parameters and computational overhead, as illustrated in Figure~\ref{fig:bdqn}. The resulting lightweight DQN model is implemented using PyTorch 2.8.0 and runs efficiently on CPU cores. \second{PELM updates the control decision every 100\,ms. The selected control parameters (CPU/GPU frequency, speculation parameters, and verification depth) are held constant for all tokens generated within this control window, and are updated only at the next control step.}

\noindent\textbf{Language Model selection.} \sys requires access to intermediate-layer outputs (e.g., via early-exit heads) that allow dynamic draft and verification depth control. In our implementation, we adopt LayerSkip~\cite{elhoushi2024layerskip}, an open-source LLM pretrained with an early-exit mechanism, as it naturally exposes such intermediate outputs. \rev{This requirement reflects a model capability rather than a model-specific dependency (see discussions in Section~\ref{sec-discussion}).}

%% file: sections/4-evaluation_setup.tex
\section{Evaluation Setup}
\begin{figure*}[h]

    \centering
    \begin{subfigure}[b]{\linewidth}
        \centering
        \includegraphics[width=\linewidth]{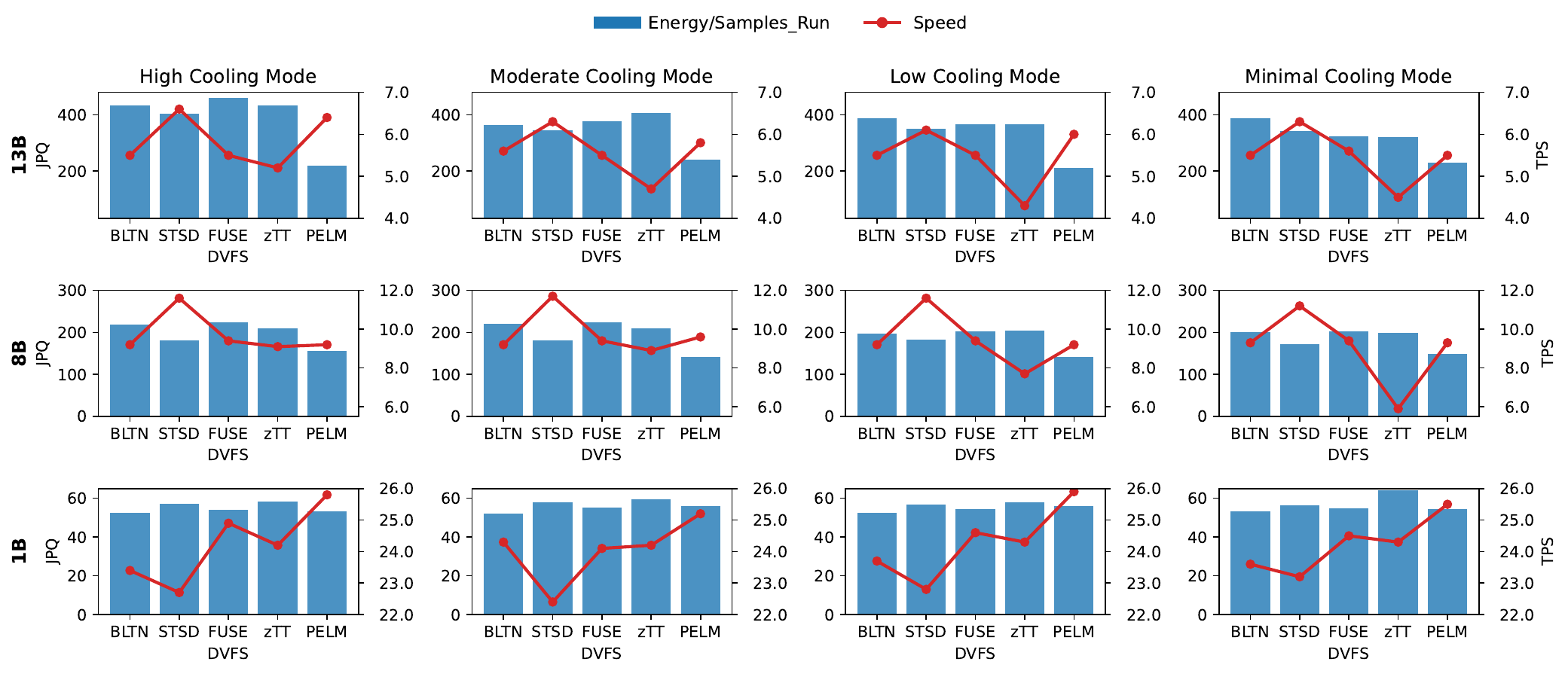}
        \caption{AGX Orin}
        \label{fig:result-main-orin}
    \end{subfigure}
    \hfill
    \begin{subfigure}[b]{\linewidth}
        \centering
        \includegraphics[width=\linewidth]{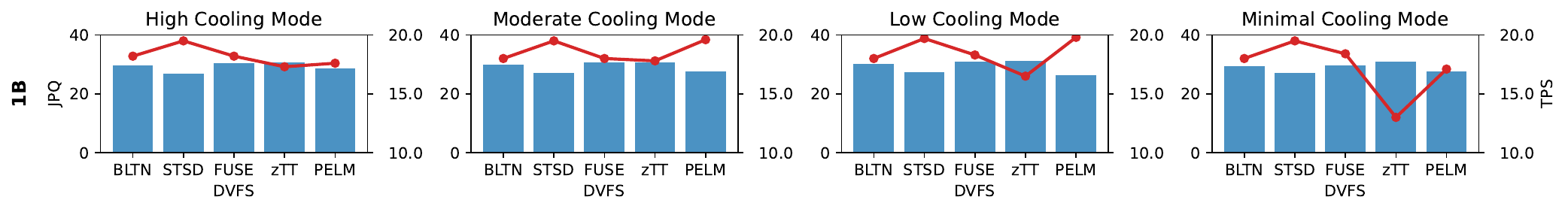}
        \caption{Orin Nano}
        \label{fig:result-main-nano}
    \end{subfigure}
    \caption{\rev{Energy consumption and decoding speed for each method across different settings. (JPQ: Joules per Query, TPS: Tokens per Second).}}
    \label{fig:result-main}
\end{figure*}

\noindent\textbf{Datasets and LLMs.} To evaluate the efficiency of our DVFS method, we construct a mobile-scale dataset by subsampling (uniform sampling with a fixed seed for reproducibility) 40 samples from each of five representative datasets which strikes a balance between evaluation comprehensiveness and the computational constraints of mobile devices: GSM8K \cite{cobbe2021training}, focusing on mathematical reasoning (\textbf{Math}); NQ-open \cite{kwiatkowski2019natural,lee-etal-2019-latent}, for question answering (\textbf{QA}); HumanEval \cite{chen2021evaluating}, for code generation (\textbf{Code}); WMT14-DE-EN \cite{bojar2014findings}, for machine translation (\textbf{Tran}); and CNN/Daily Mail \cite{hermann2015teaching}, for text summarization (\textbf{Sum}). This selection ensures a diverse range of domains for evaluation. For LLMs, we select LLaMA-3.2-1B, LLaMA-3.1-8B, and LLaMA-2-13B models pretrained using LayerSkip \cite{elhoushi2024layerskip} for evaluation (Section~\ref{sec-method-implementation}), covering different sizes that allow us to evaluate the DVFS governor under distinct computational workloads and thermal stress. These models are executed using the HuggingFace Transformers library \cite{wolf-etal-2020-transformers} with a PyTorch backend. \add{The decoding is configured with a batch size of one, approximating the single-request behavior of on-device mobile queries.}

\noindent\textbf{Test-bed.} We evaluate our DVFS governor on two mobile/embedded platforms: the \textbf{NVIDIA Jetson AGX Orin} and \textbf{Orin Nano} (details in Section \ref{sec-method-implementation}). To evaluate robustness under varying thermal conditions, we define four cooling conditions by fixing fan PWM levels: 64 (High), 51 (Moderate), 38 (Low), and 26 (Minimum) for the AGX Orin. Due to the Nano's limited heat dissipation, its levels are set higher: 90 (High), 77 (Moderate), 64 (Low), and 51 (Minimum). To ensure a fair comparison from a consistent thermal baseline, all sessions are initiated at 45$^\circ$C. We define thermal failure thresholds at 78$^\circ$C (AGX Orin) and 73$^\circ$C (Nano); any session reaching its limit is forcibly terminated, marking a failure in the \textbf{CR} metric. The DVFS "warning" temperature is set 5$^\circ$C below this critical point. Finally, we set the target decoding speed for our governor to match the default system's baseline performance: (AGX Orin) 25 Tokens/s (1B), 10 Tokens/s (8B), and 6 Tokens/s (13B); and (Nano) 18 Tokens/s (1B).

\noindent\textbf{Evaluation Metrics.} We evaluate our method using two categories of metrics: \textbf{(1) DVFS-Level Efficiency} and \textbf{(2) Task-Level Quality}. Our \textbf{DVFS-Level Efficiency} metrics, the primary focus of this work, quantify the runtime performance of our governor. These include: \textbf{Decoding Speed} (Tokens/s); \textbf{Energy Consumption} (Joules), the total energy consumed per session; \textbf{Completion Rate (CR)}, the percentage of samples successfully completed;  \textbf{Layers Per Accepted Token (LPAT)}, the average number of layers executed per accepted token; \textbf{Mean Variable Depth (VDm)}; and \textbf{Speculation Ratio (SR)}, which measures the speedup from our speculation method.

For \textbf{Task-Level Quality}, we verify that efficiency gains do not lead to catastrophic degradation in output quality by reporting standard \rev{task-specific metrics (\textbf{Accuracy, EM, pass@1, BLEU, and ROUGE-L}) separately for each benchmark (Table~\ref{tab:task-performance}). These metrics directly capture correctness and generation quality, allowing us to detect potential degradation under variable-depth execution.} For a comprehensive comparison, we additionally normalize each task score to a 0–100 scale and average them to compute the \textbf{Task Performance Average (TSA)}, which serves as a summary indicator. \rev{Finally, we define \textbf{PPJ} (Performance per joule, values are scaled by $10^3$ for readability.) as $\text{TSA} / \text{Energy (J)}$ to quantify the joint efficiency–quality trade-off.}

\noindent\textbf{Baseline methods.} We compare \sys~with the following baseline methods:

\begin{itemize}
    \item \textbf{Built-in \rev{(BLTN)}.} The system’s default DVFS governors -- \textit{schedutil} for the CPU and \textit{nvhost\_podgov} for the GPU.
    \rev{\item \textbf{Standard Self-Speculative Decoding (STSD).} A static speculative decoding baseline, in which the first quarter of the model functions as the draft model, while the complete model acts as the target model. The draft length is fixed at 3. The default system DVFS governor is used throughout STSD.}
    \item \textbf{FUSE} \cite{zhang2025dissecting}. A DVFS governor for mobile LLMs. It uses a search-based approach to find optimal frequencies that either minimize energy for a target speed or maximize speed for an energy budget.
    \item \textbf{zTT} \cite{kim2022ztt}. An RL-based DVFS framework for DNN tasks. Its primary goal is to learn a policy that maximizes performance while actively mitigating thermal throttling to maintain QoE.
\end{itemize}

In the next section, we demonstrate how \sys's co-design of DVFS and speculative decoding outperforms the system built-in governor and conventional QoE-based methods.

%% file: sections/5-results.tex
\section{Results}

\subsection{Power Governing Performance}
\label{sec-results-dvfs-performance}

\begin{figure}[htbp]
    \centering
    \begin{subfigure}[b]{0.5\columnwidth}
        \centering
        \includegraphics[width=0.9\linewidth]{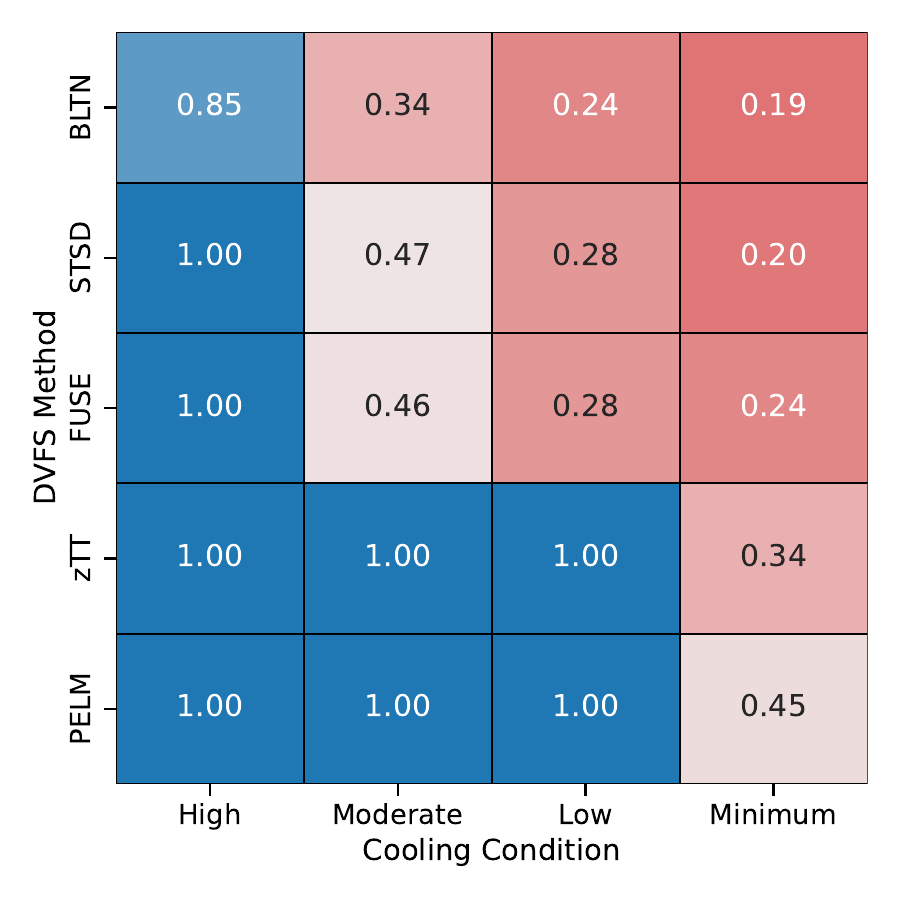}
        \caption{AGX Orin -- 13B}
        \label{fig:main-robust-o13b}
    \end{subfigure}%
    \hfill
    \begin{subfigure}[b]{0.5\columnwidth}
        \centering
        \includegraphics[width=0.9\columnwidth]{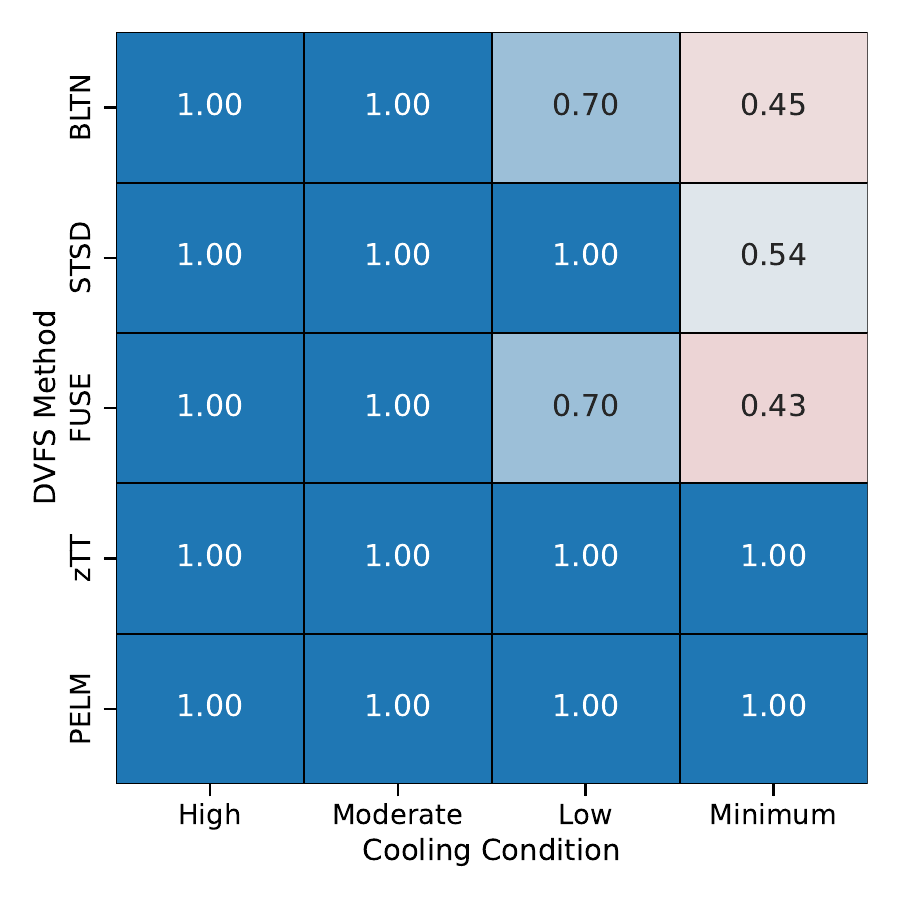}
        \caption{AGX Orin -- 8B}
        \label{fig:main-robust-o8b}
    \end{subfigure}
    
    \centering
    \begin{subfigure}[b]{0.5\columnwidth}
        \centering
        \includegraphics[width=0.9\linewidth]{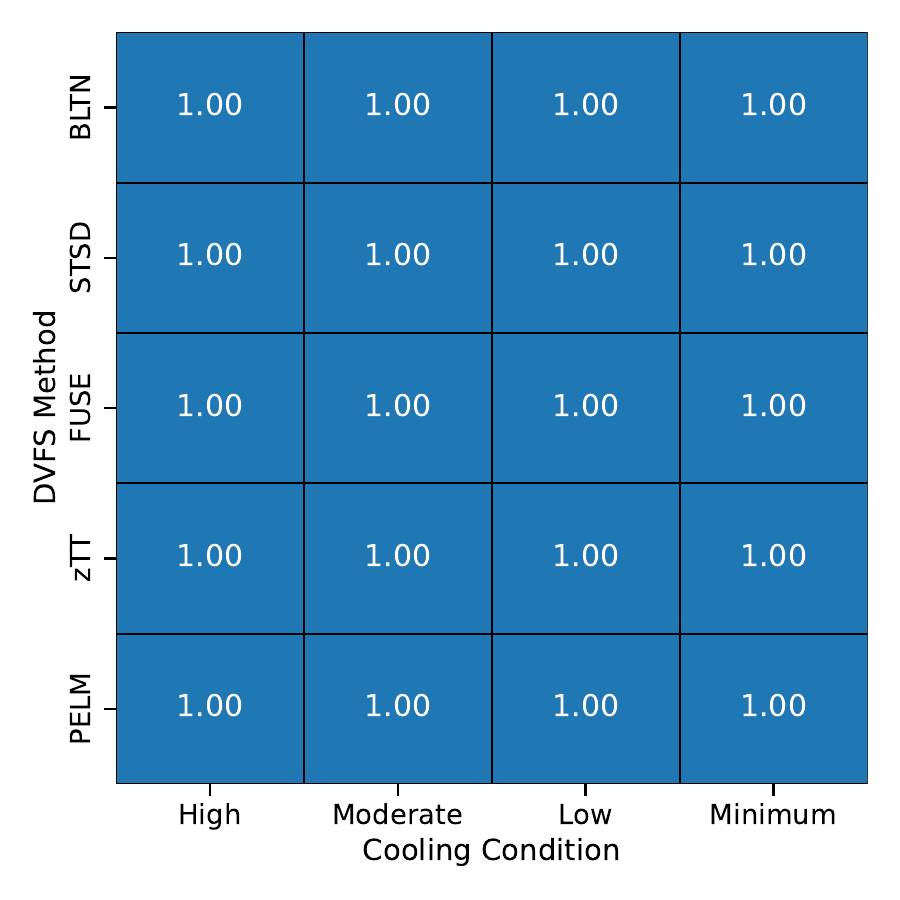}
        \caption{AGX Orin -- 1B}
        \label{fig:main-robust-o1b}
    \end{subfigure}%
    \hfill
    \begin{subfigure}[b]{0.5\columnwidth}
        \centering
        \includegraphics[width=0.9\columnwidth]{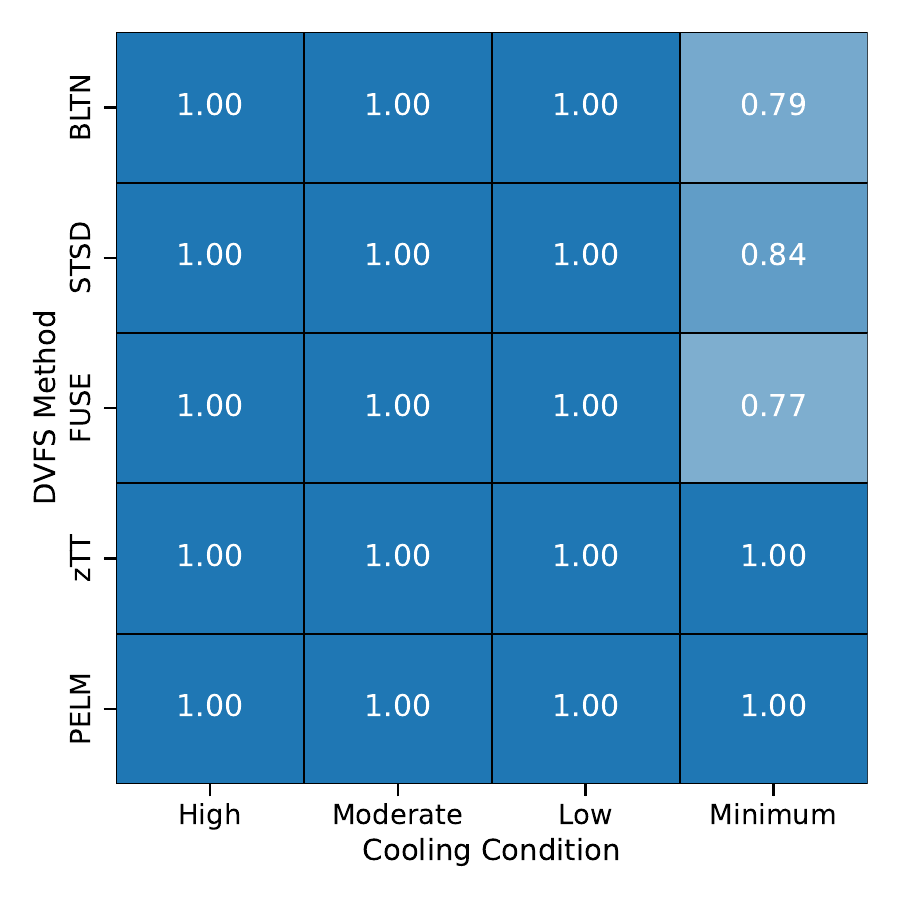}
        \caption{Orin Nano -- 1B}
        \label{fig:main-robust-n1b}
    \end{subfigure}
    \caption{\rev{Completion Rate (CR) for different DVFS under diverse thermal conditions.}}
    \label{fig:main-robust}
\end{figure}

\begin{table}
\centering
\caption{\rev{Impact of DVFS Strategies on GPU/CPU Average Frequency Ratios on Jetson Platforms.}}
\label{tab:frequency}
\resizebox{\columnwidth}{!}{
\begin{tabular}{lllcccccccc}
\toprule
\multirow{3}{*}{\textbf{H}} & \multirow{3}{*}{\textbf{\#P}} & \multirow{3}{*}{\textbf{DVFS}} & \multicolumn{8}{c}{\textbf{Avg. Freq Ratio}} \\
\cmidrule(lr){4-11}
& & & \multicolumn{2}{c}{\textbf{High}} & \multicolumn{2}{c}{\textbf{Moderate}} & \multicolumn{2}{c}{\textbf{Low}} & \multicolumn{2}{c}{\textbf{Minimum}} \\
\cmidrule(lr){4-5} \cmidrule(lr){6-7} \cmidrule(lr){8-9} \cmidrule(lr){10-11}
& & & \textbf{GPU} & \textbf{CPU} & \textbf{GPU} & \textbf{CPU} & \textbf{GPU} & \textbf{CPU} & \textbf{GPU} & \textbf{CPU} \\
\midrule

\multirow{12}{*}{\rotatebox{90}{\textbf{AGX Orin}}} & \multirow{5}{*}{\textbf{13B}} & BLTN & 0.91 & 0.33 & 0.91 & 0.35 & 0.91 & 0.33 & 0.90 & 0.33 \\
& & STSD &  0.91 & 0.33 & 0.91 & 0.33 & 0.91 & 0.33 & 0.91 & 0.33 \\
& & FUSE & 0.83 & 1.00 & 0.83 & 1.00 & 0.83 & 1.00 & 0.83 & 1.00 \\
& & zTT & 0.74 & 0.77 & 0.63 & 0.58 & 0.47 & 0.38 & 0.55 & 0.50 \\
& & PELM & 0.57 & 0.72 & 0.47 & 0.63 & 0.49 & 0.65 & 0.48 & 0.59 \\
\cmidrule(lr){2-11}
& \multirow{5}{*}{\textbf{8B}} & BLTN & 0.74 & 0.34 & 0.74 & 0.34 & 0.72 & 0.35 & 0.72 & 0.34 \\
& & STSD & 0.68 & 0.49 & 0.68 & 0.49 & 0.68 & 0.48 & 0.67 & 0.49 \\
& & FUSE & 0.92 & 1.00 & 0.92 & 1.00 & 0.92 & 1.00 & 0.92 & 1.00 \\
& & zTT & 0.83 & 0.92 & 0.80 & 0.89 & 0.71 & 0.69 & 0.53 & 0.44 \\
& & PELM & 0.62 & 0.73 & 0.53 & 0.72 & 0.50 & 0.65 & 0.40 & 0.64 \\
\cmidrule(lr){2-11}
& \multirow{5}{*}{\textbf{1B}} & BLTN & 0.28 & 0.35 & 0.29 & 0.35 & 0.29 & 0.35 & 0.28 & 0.36 \\
& & STSD & 0.29 & 0.50 & 0.28 & 0.50 & 0.30 & 0.50 & 0.30 & 0.49 \\
& & FUSE & 0.43 & 1.00 & 0.43 & 1.00 & 0.43 & 1.00 & 0.43 & 1.00 \\
& & zTT & 0.61 & 0.98 & 0.64 & 0.98 & 0.60 & 0.98 & 0.77 & 0.97 \\
& & PELM & 0.52 & 0.70 & 0.60 & 0.75 & 0.57 & 0.70 & 0.59 & 0.68 \\
\midrule

\multirow{5}{*}{\rotatebox{90}{\textbf{Orin Nano}}} & \multirow{4}{*}{\textbf{1B}} & BLTN & 0.38 & 0.39 & 0.38 & 0.39 & 0.38 & 0.39 & 0.37 & 0.39 \\
& & STSD & 0.34 & 0.38 & 0.34 & 0.38 & 0.34 & 0.38 & 0.33 & 0.38 \\
& & FUSE & 0.50 & 0.76 & 0.50 & 0.76 & 0.50 & 0.76 & 0.50 & 0.76 \\
& & zTT & 0.45 & 0.74 & 0.51 & 0.73 & 0.53 & 0.68 & 0.48 & 0.54 \\
& & PELM & 0.53 & 0.66 & 0.49 & 0.64 & 0.48 & 0.65 & 0.49 & 0.64 \\
\bottomrule
\end{tabular}
}
\end{table}

\noindent\textbf{Overall performance.} We first compare \sys~with baseline methods in terms of two primary metrics to evaluate their overall performance: energy consumed per query and token decoding speed. As shown in Figure~\ref{fig:result-main-orin}, for LLaMA-13B model running on AGX Orin, \sys~has the least energy consumption across all cooling conditions, \rev{achieving 29.0\% to 52.4\% energy savings compared to the other four methods, while keeping the highest average generation speed under three out of four cooling conditions. On 8B models, \sys~consumes 13.4\% to 32.5\% less energy per query than baselines, with comparable generation speed}.

\rev{For the 1B model on AGX Orin, \sys~consumes slightly higher energy (0.4\%–7.3\% above the lowest-energy baseline) but consistently meets the 25 tokens/s target, which several frequency-only DVFS methods fail to achieve. This suggests that while hardware-level frequency scaling remains effective under light workloads with sufficient thermal headroom, its optimization space is limited. On Orin Nano, where the same 1B model represents a relatively heavier workload for the device, \sys~achieves low energy consumption while maintaining competitive speed across cooling modes, demonstrating stable performance under tighter resource constraints.}

\rev{Overall, \sys's advantage becomes increasingly pronounced as workload intensity grows. Frequency-only governors are bounded by hardware-level tuning and tend to either over-prioritize performance (e.g., BLTN, STSD, FUSE) or conservatively react to thermal throttling (e.g., zTT). By jointly controlling frequency and decoding behavior, \sys~consistently achieves the best or comparable performance across thermal conditions while improving energy efficiency.}

\noindent\textbf{Robustness under different thermal conditions.} We evaluate the DVFS thermal robustness by checking the query completion rate (CR) for each case. In Figure~\ref{fig:main-robust}, the results clearly divide the methods into two groups: those with thermal awareness and those without. \rev{Methods without thermal awareness (e.g., BLTN and FUSE) frequently fail to complete the entire query set, with CR dropping as low as 19\% in the 13B workload. Although STSD achieves superior token throughput in many scenarios, it underperforms with respect to this critical metric.} In contrast, both zTT and \sys demonstrate exceptional robustness, achieving 100\% completion rate in all scenarios except one. The single failure case --- the extreme 13B workload on minimum cooling --- serves as a tie-breaker. In this worst-case scenario, \sys~completed 45\% of queries before hitting the limit, while zTT completed only 34\%. This means \sys~processed 32.4\% more queries than the closest competitor, proving its superior robustness when the system is under maximum thermal stress.

\noindent\textbf{Adaptive Frequency Scaling Behavior.} \add{To investigate the mechanisms behind the optimized power and speed, we evaluated the average operating frequency ratios (summarized in Table~\ref{tab:frequency}). The results demonstrate \sys's adaptive behavior, which manifests differently under heavy and light workloads.}

\add{For heavy and moderate workloads (13B and 8B models on AGX Orin), \sys consistently operates at the lowest GPU frequency among all methods. This directly explains the previously observed energy savings. \rev{For instance, in the 13B-High scenario, \sys (PELM) runs at 0.57, which is 37.4\% lower than BLTN and STSD (0.91) and 31.3\% lower than FUSE (0.83)}. As thermal pressure increases (e.g., in Low or Minimum cooling), \sys adaptively lowers its frequency further, whereas BLTN and FUSE maintain their high, inefficient frequencies.}

\add{\rev{Conversely, for light workloads (1B model on both platforms), \sys shows a critical adaptation for performance. On Jetson Orin, it operates at a higher GPU frequency (e.g., 0.52 in High mode) compared to the overly conservative BLTN (0.28) and FUSE (0.43) governors while avoiding a substantial increase in power consumption}. This finding is crucial: \sys correctly identifies the ample thermal headroom and proactively increases its frequency to meet the QoE (token generation speed) target. This directly explains the "Overall Performance" result where \sys was the only method to meet the target speed, while others failed precisely because their frequencies were too low.}

\add{This intelligent, bi-directional adaptation (saving power when heavy, boosting performance when light) is also observed on the Jetson Orin Nano. These results demonstrate that \sys possesses a genuine awareness of both workload demands and thermal conditions, allowing it to adapt its strategy to optimize for either energy efficiency or performance as needed.}

\subsection{Task Performance}

\begin{table}
\centering
\caption{\rev{Task score comparison between \sys~and the baselines. (TSA=Task Score Average, PPJ=Performance per joule)}}
\label{tab:task-performance}
\resizebox{0.85\columnwidth}{!}{
\begin{tabular}{ccccccccccc}
\toprule
\textbf{H} & \textbf{\#P} & \textbf{Cooling} & \textbf{DVFS} & \textbf{Math} & \textbf{QA} & \textbf{Code} & \textbf{Tran} & \textbf{Sum} & \textbf{TSA} & \textbf{PPJ} \\
\midrule
\multirow{48}{*}{\rotatebox{90}{\textbf{Jetson AGX Orin}}} & \multirow{16}{*}{\textbf{13B}} & \multirow{5}{*}{High} & BLTN & 2.94 & 30.00 & 0.00 & 21.51 & 26.50 & 16.19 & 37.30 \\
 & & & FUSE & 2.50 & 28.75 & 0.00 & 22.73 & 25.96 & 15.99 & 34.75 \\
 & & & STSD & 5.00 & 28.75 & 0.00 & 22.71 & 25.96 & 16.49 & 40.65 \\
 & & & zTT & 2.50 & 28.75 & 0.00 & 22.73 & 25.96 & 15.99 & 36.92 \\
 & & & PELM & 5.00 & 15.00 & 0.00 & 7.52 & 26.12 & 10.73 & 48.82 \\
\cmidrule{3-11}
 & & \multirow{5}{*}{Moderate} & BLTN & 9.09 & 20.00 & 0.00 & 23.97 & 27.71 & 16.15 & 44.48 \\
 & & & STSD & 7.14 & 30.95 & 0.00 & 22.08 & 27.37 & 17.51 & 50.76 \\
 & & & FUSE & 7.69 & 30.95 & 0.00 & 22.72 & 26.68 & 17.61 & 46.63 \\
 & & & zTT & 2.50 & 28.75 & 0.00 & 22.73 & 25.96 & 15.99 & 39.41 \\
 & & & PELM & 10.00 & 12.50 & 0.00 & 9.68 & 26.92 & 11.82 & 49.32 \\
\cmidrule{3-11}
 & & \multirow{5}{*}{Low} & BLTN & 0.00 & 18.18 & 0.00 & 26.67 & 30.03 & 14.98 & 38.60 \\
 & & & STSD & 0.00 & 18.18 & 0.00 & 27.38 & 29.25 & 14.96 & 42.68 \\
 & & & FUSE & 0.00 & 18.18 & 0.00 & 27.03 & 29.25 & 14.89 & 40.66 \\
 & & & zTT & 2.50 & 28.75 & 0.00 & 22.73 & 25.96 & 15.99 & 43.44 \\
 & & & PELM & 5.00 & 10.00 & 0.00 & 6.59 & 26.41 & 9.60 & 45.39 \\
\cmidrule{3-11}
 & & \multirow{5}{*}{Minimum} & BLTN & 0.00 & 12.50 & 0.00 & 26.17 & 31.33 & 14.00 & 36.02 \\
 & & & STSD & 0.00 & 10.00 & 0.00 & 26.17 & 31.33 & 13.50 & 39.56 \\
 & & & FUSE & 0.00 & 18.18 & 0.00 & 26.67 & 31.33 & 15.23 & 46.95 \\
 & & & zTT & 9.09 & 20.00 & 0.00 & 23.97 & 27.71 & 16.15 & 50.23 \\
 & & & PELM & 7.69 & 16.67 & 0.00 & 10.52 & 25.09 & 11.99 & 52.42 \\
\cmidrule{2-11}
 & \multirow{16}{*}{\textbf{8B}} & \multirow{5}{*}{High} & BLTN & 20.00 & 22.50 & 27.50 & 18.76 & 21.75 & 22.10 & 100.72 \\
 & & & STSD & 22.50 & 23.75 & 27.50 & 18.76 & 21.44 & 22.79 & 126.19 \\
 & & & FUSE & 20.00 & 22.50 & 27.50 & 18.76 & 21.75 & 22.10 & 98.93 \\
 & & & zTT & 20.00 & 22.50 & 27.50 & 18.76 & 21.75 & 22.10 & 105.05 \\
 & & & PELM & 15.00 & 22.50 & 25.00 & 18.51 & 22.57 & 20.71 & 133.26 \\
\cmidrule{3-11}
 & & \multirow{5}{*}{Moderate} & BLTN & 20.00 & 22.50 & 27.50 & 18.76 & 21.75 & 22.10 & 100.22 \\
 & & & STSD & 22.50 & 23.75 & 27.50 & 18.76 & 21.44 & 22.79 & 125.63 \\
 & & & FUSE & 20.00 & 22.50 & 27.50 & 18.76 & 21.75 & 22.10 & 98.47 \\
 & & & zTT & 20.00 & 22.50 & 27.50 & 18.76 & 21.75 & 22.10 & 105.43 \\
 & & & PELM & 27.50 & 22.50 & 25.00 & 14.61 & 23.83 & 22.69 & 160.91 \\
\cmidrule{3-11}
 & & \multirow{5}{*}{Low} & BLTN & 23.33 & 23.33 & 23.33 & 14.49 & 22.54 & 21.41 & 108.32 \\
 & & & STSD & 22.50 & 23.75 & 27.50 & 18.76 & 21.44 & 22.79 & 124.87 \\
 & & & FUSE & 23.33 & 23.33 & 23.33 & 14.49 & 22.54 & 21.41 & 105.80 \\
 & & & zTT & 20.00 & 22.50 & 27.50 & 18.76 & 21.75 & 22.10 & 108.63 \\
 & & & PELM & 25.00 & 18.75 & 27.50 & 12.23 & 23.33 & 21.36 & 149.90 \\
\cmidrule{3-11}
 & & \multirow{5}{*}{Minimum} & BLTN & 30.77 & 21.43 & 27.27 & 16.43 & 21.23 & 23.43 & 117.18 \\
 & & & STSD & 25.00 & 19.57 & 26.09 & 15.76 & 22.45 & 21.77 & 126.09 \\
 & & & FUSE & 30.77 & 20.00 & 28.57 & 17.01 & 21.35 & 23.54 & 116.01 \\
 & & & zTT & 20.00 & 22.50 & 27.50 & 18.76 & 21.75 & 22.10 & 110.66 \\
 & & & PELM & 25.00 & 17.50 & 20.00 & 11.29 & 22.20 & 19.20 & 128.99 \\
\cmidrule{2-11}
 & \multirow{16}{*}{\textbf{1B}} & \multirow{5}{*}{High} & BLTN & 2.50 & 3.75 & 7.50 & 10.21 & 22.10 & 9.21 & 175.90 \\
 & & & STSD & 2.50 & 3.75 & 7.50 & 10.35 & 21.53 & 9.13 & 159.89 \\
 & & & FUSE & 2.50 & 3.75 & 7.50 & 10.21 & 22.10 & 9.21 & 170.52 \\
 & & & zTT & 2.50 & 3.75 & 7.50 & 10.21 & 22.10 & 9.21 & 157.82 \\
 & & & PELM & 0.00 & 3.75 & 10.00 & 9.36 & 20.19 & 8.66 & 162.11 \\
\cmidrule{3-11}
 & & \multirow{5}{*}{Moderate} & BLTN & 2.50 & 3.75 & 7.50 & 10.21 & 22.10 & 9.21 & 177.45 \\
 & & & STSD & 2.50 & 3.75 & 7.50 & 10.35 & 21.53 & 9.13 & 158.13 \\
 & & & FUSE & 2.50 & 3.75 & 7.50 & 10.21 & 22.10 & 9.21 & 167.45 \\
 & & & zTT & 2.50 & 3.75 & 7.50 & 10.21 & 22.10 & 9.21 & 154.49 \\
 & & & PELM & 2.50 & 3.75 & 7.50 & 9.50 & 21.34 & 8.92 & 159.89 \\
\cmidrule{3-11}
 & & \multirow{5}{*}{Low} & BLTN & 2.50 & 3.75 & 7.50 & 10.21 & 22.10 & 9.21 & 175.00 \\
 & & & STSD & 2.50 & 3.75 & 7.50 & 10.35 & 21.53 & 9.13 & 160.61 \\
 & & & FUSE & 2.50 & 3.75 & 7.50 & 10.21 & 22.10 & 9.21 & 168.75 \\
 & & & zTT & 2.50 & 3.75 & 7.50 & 10.21 & 22.10 & 9.21 & 158.60 \\
 & & & PELM & 2.50 & 3.75 & 7.50 & 9.99 & 21.92 & 9.13 & 163.27 \\
\cmidrule{3-11}
 & & \multirow{5}{*}{Minimum} & BLTN & 2.50 & 3.75 & 7.50 & 10.21 & 22.10 & 9.21 & 172.85 \\
 & & & STSD & 2.50 & 3.75 & 7.50 & 10.35 & 21.53 & 9.13 & 161.78 \\
 & & & FUSE & 2.50 & 3.75 & 7.50 & 10.21 & 22.10 & 9.21 & 168.48 \\
 & & & zTT & 2.50 & 3.75 & 7.50 & 10.21 & 22.10 & 9.21 & 143.90 \\
 & & & PELM & 2.50 & 3.75 & 7.50 & 9.42 & 21.13 & 8.86 & 163.12 \\
\midrule
\multirow{16}{*}{\rotatebox{90}{\textbf{Jetson Orin Nano}}} & \multirow{16}{*}{\textbf{1B}} & \multirow{5}{*}{High} & BLTN & 2.50 & 3.75 & 7.50 & 10.21 & 22.10 & 9.21 & 309.82 \\
 & & & STSD & 2.50 & 3.75 & 7.50 & 10.21 & 21.48 & 9.09 & 337.87 \\
 & & & FUSE & 2.50 & 2.50 & 7.50 & 8.49 & 21.54 & 8.51 & 279.44 \\
 & & & zTT & 2.50 & 3.75 & 7.50 & 10.21 & 22.10 & 9.21 & 300.97 \\
 & & & PELM & 2.50 & 2.50 & 7.50 & 8.49 & 21.54 & 8.51 & 297.56 \\
\cmidrule{3-11}
 & & \multirow{5}{*}{Moderate} & BLTN & 2.50 & 3.75 & 7.50 & 10.21 & 22.10 & 9.21 & 307.54 \\
 & & & STSD & 2.50 & 3.75 & 7.50 & 10.21 & 21.48 & 9.09 & 334.87 \\
 & & & FUSE & 5.00 & 2.50 & 5.00 & 5.93 & 20.58 & 7.80 & 253.92 \\
 & & & zTT & 2.50 & 3.75 & 7.50 & 10.21 & 22.10 & 9.21 & 301.37 \\
 & & & PELM & 5.00 & 2.50 & 5.00 & 5.93 & 20.58 & 7.80 & 283.03 \\
\cmidrule{3-11}
 & & \multirow{5}{*}{Low} & BLTN & 2.50 & 3.75 & 7.50 & 10.21 & 22.10 & 9.21 & 305.44 \\
 & & & STSD & 2.50 & 3.75 & 7.50 & 10.21 & 21.48 & 9.09 & 331.98 \\
 & & & FUSE & 2.50 & 3.75 & 7.50 & 10.21 & 22.10 & 9.21 & 298.33 \\
 & & & zTT & 2.50 & 3.75 & 7.50 & 6.51 & 21.13 & 8.28 & 264.39 \\
 & & & PELM & 2.50 & 3.75 & 7.50 & 6.51 & 21.13 & 8.28 & 315.67 \\
\cmidrule{3-11}
 & & \multirow{5}{*}{Minimum} & BLTN & 2.50 & 3.75 & 7.50 & 10.21 & 22.10 & 9.21 & 313.31 \\
 & & & STSD & 2.94 & 4.29 & 9.68 & 10.42 & 21.94 & 9.85 & 363.56 \\
 & & & FUSE & 0.00 & 3.75 & 10.00 & 7.38 & 20.58 & 8.34 & 280.62 \\
 & & & zTT & 3.03 & 4.55 & 9.68 & 10.45 & 22.64 & 10.07 & 326.03 \\
 & & & PELM & 0.00 & 3.75 & 10.00 & 7.38 & 20.58 & 8.34 & 302.58 \\
\bottomrule
\end{tabular}
}
\end{table}
The variable depth execution in \sys, which dynamically uses a partial LLM, could potentially impact downstream task performance. We now evaluate this impact comprehensively. We compare \sys's task scores against the vanilla (auto-regressive) decoding used by the baseline methods. The results are shown in Table~\ref{tab:task-performance}. 

The primary concern is: does speculative decoding with variable depth execution degrade the task performance? As shown in Table~\ref{tab:task-performance}, for 8B and 1B models, \rev{the overall task scores (TSA) remain statistically comparable to those of the vanilla full-model decoding on both Jetson platforms, with the largest degradation being less than 17.2\%.}  For the 13B model, the output still retains meaningful quality to full depth vanilla baseline. These results show the effectiveness of variable depth execution without sacrificing task performance.

\rev{Now we turn to performance efficiency. The \textbf{PPJ} column quantifies output quality normalized by energy consumption.} In all moderate-to-heavy workloads (13B, 8B@Orin), \sys is significantly more efficient, delivering responses of comparable quality with a lower energy cost at target speed -- achieving up to 45.4\% higher performance-per-joule than baselines.

Finally, the 1B results on both AGX Orin and Orin Nano may appear different at first glance, as \sys~does not always achieve the highest PPJ among all baselines. This behavior is expected and consistent with the trends observed in Figure~\ref{fig:result-main} and Figure~\ref{fig:main-robust}. Under light workloads with minimal thermal pressure, the optimization objective shifts from aggressive energy minimization to QoE-aware performance control. In this regime, \sys intentionally allocates slightly more power to reliably meet the target decoding speed (25 tokens/s) while maintaining competitive energy efficiency. On Orin Nano, due to its more constrained compute capability, \sys operates under a tighter power–performance envelope and adopts a more balanced tradeoff between power and speed, whereas several ``more efficient'' baselines reduce power but fail to consistently sustain the desired throughput. These results demonstrate that \sys remains effective even in low-demand scenarios by adaptively balancing speed and efficiency, rather than over-optimizing for a single objective.


\subsection{Ablations}
\label{sec-results-ablation}

\begin{table}[h]
\centering
\caption{\rev{Ablation study of \sys components. (TSA=Task Score Average, SR=Speculation Ratio, PPJ=Performance per joule)}}
\label{tab:ablation_study_components}

\resizebox{\columnwidth}{!}{
\begin{tabular}{ccccccccccccccc}
\toprule
\textbf{H} & \textbf{\#P} & \textbf{Cooling} & \textbf{DVFS} & \textbf{VD} & \textbf{SSD} & \textbf{FC} & \textbf{CR} & \textbf{Energy} & \textbf{Speed} & \textbf{LPATm} & \textbf{VDm} & \textbf{SR} & \textbf{TSA} & \textbf{PPJ} \\
\midrule
\multirow{48}{*}{\rotatebox{90}{\textbf{Jetson AGX Orin}}} & \multirow{16}{*}{\textbf{13B}} & \multirow{4}{*}{High} & PELM & Y & Y & Y & 1.00 & 43724 & 6.40 & 31.00 & 31.77 & 1.02 & 10.73 & 48.82 \\
 & & & PELM\_A1 & N & Y & Y & 1.00 & 66247 & 5.60 & 35.97 & 40.00 & 1.11 & 16.06 & 48.23 \\
 & & & PELM\_A2 & Y & N & Y & 1.00 & 52732 & 6.10 & 32.63 & 32.64 & 1.00 & 10.23 & 38.59 \\
 & & & PELM\_A3 & Y & Y & N & 1.00 & 59446 & 6.80 & 32.39 & 33.18 & 1.02 & 11.07 & 37.05 \\
\cmidrule{3-15}
 & & \multirow{4}{*}{Moderate} & PELM & Y & Y & Y & 1.00 & 47696 & 5.80 & 31.95 & 32.72 & 1.02 & 11.82 & 49.32 \\
 & & & PELM\_A1 & N & Y & Y & 1.00 & 71966 & 5.30 & 37.85 & 40.00 & 1.05 & 16.07 & 44.43 \\
 & & & PELM\_A2 & Y & N & Y & 1.00 & 44803 & 6.20 & 31.54 & 31.53 & 1.00 & 9.86 & 43.80 \\
 & & & PELM\_A3 & Y & Y & N & 0.53 & 27809 & 7.00 & 31.76 & 33.02 & 1.04 & 9.13 & 34.82 \\
\cmidrule{3-15}
 & & \multirow{4}{*}{Low} & PELM & Y & Y & Y & 1.00 & 42084 & 6.00 & 30.76 & 31.58 & 1.03 & 9.60 & 45.39 \\
 & & & PELM\_A1 & N & Y & Y & 0.71 & 39957 & 5.20 & 36.77 & 40.00 & 1.09 & 16.23 & 57.26 \\
 & & & PELM\_A2 & Y & N & Y & 0.71 & 30603 & 5.90 & 32.40 & 32.40 & 1.00 & 9.23 & 42.52 \\
 & & & PELM\_A3 & Y & Y & N & 0.33 & 17785 & 6.90 & 32.34 & 34.10 & 1.05 & 12.02 & 44.59 \\
\cmidrule{3-15}
 & & \multirow{4}{*}{Minimum} & PELM & Y & Y & Y & 0.45 & 20363 & 5.50 & 33.49 & 34.29 & 1.02 & 11.00 & 52.42 \\
 & & & PELM\_A1 & N & Y & Y & 0.37 & 20283 & 4.90 & 39.09 & 40.00 & 1.02 & 17.01 & 62.07 \\
 & & & PELM\_A2 & Y & N & Y & 0.45 & 17720 & 5.90 & 32.73 & 32.74 & 1.00 & 8.66 & 43.48 \\
 & & & PELM\_A3 & Y & Y & N & 0.23 & 14207 & 6.90 & 32.49 & 34.51 & 1.06 & 13.76 & 44.54 \\
\cmidrule{2-15}
 & \multirow{16}{*}{\textbf{8B}} & \multirow{4}{*}{High} & PELM & Y & Y & Y & 1.00 & 30934 & 9.20 & 26.07 & 29.62 & 1.14 & 20.71 & 133.26 \\
 & & & PELM\_A1 & N & Y & Y & 1.00 & 32237 & 9.40 & 27.28 & 32.00 & 1.18 & 22.60 & 139.51 \\
 & & & PELM\_A2 & Y & N & Y & 1.00 & 35151 & 9.00 & 28.93 & 28.88 & 1.00 & 22.82 & 129.22 \\
 & & & PELM\_A3 & Y & Y & N & 1.00 & 35311 & 11.00 & 27.29 & 29.85 & 1.10 & 21.97 & 123.80 \\
\cmidrule{3-15}
 & & \multirow{4}{*}{Moderate} & PELM & Y & Y & Y & 1.00 & 28059 & 9.60 & 24.84 & 28.55 & 1.15 & 22.69 & 160.91 \\
 & & & PELM\_A1 & N & Y & Y & 1.00 & 30500 & 9.10 & 26.54 & 32.00 & 1.20 & 22.89 & 149.34 \\
 & & & PELM\_A2 & Y & N & Y & 1.00 & 31680 & 9.60 & 27.05 & 27.04 & 1.00 & 20.26 & 127.23 \\
 & & & PELM\_A3 & Y & Y & N & 1.00 & 35705 & 11.60 & 25.61 & 27.80 & 1.09 & 21.04 & 117.29 \\
 \cmidrule{3-15}
 & & \multirow{4}{*}{Low} & PELM & Y & Y & Y & 1.00 & 28360 & 9.20 & 23.95 & 26.93 & 1.12 & 21.36 & 149.90 \\
 & & & PELM\_A1 & N & Y & Y & 1.00 & 31752 & 8.60 & 26.87 & 32.00 & 1.19 & 22.78 & 142.79 \\
 & & & PELM\_A2 & Y & N & Y & 1.00 & 30829 & 9.60 & 25.99 & 25.99 & 1.00 & 17.20 & 111.03 \\
 & & & PELM\_A3 & Y & Y & N & 0.86 & 28554 & 11.50 & 25.73 & 28.27 & 1.10 & 20.35 & 122.59 \\
\cmidrule{3-15}
 & & \multirow{4}{*}{Minimum} & PELM & Y & Y & Y & 1.00 & 29617 & 9.30 & 23.32 & 26.86 & 1.15 & 19.20 & 128.99 \\
 & & & PELM\_A1 & N & Y & Y & 0.96 & 33071 & 8.00 & 27.55 & 32.00 & 1.16 & 22.82 & 133.16 \\
 & & & PELM\_A2 & Y & N & Y & 0.83 & 24233 & 9.40 & 26.36 & 26.34 & 1.00 & 16.85 & 115.43 \\
 & & & PELM\_A3 & Y & Y & N & 0.53 & 17695 & 11.50 & 25.93 & 28.29 & 1.09 & 21.22 & 127.12 \\
\cmidrule{2-15}
 & \multirow{16}{*}{\textbf{1B}} & \multirow{4}{*}{High} & PELM & Y & Y & Y & 1.00 & 10629 & 25.80 & 12.95 & 14.89 & 1.15 & 8.66 & 162.11 \\
 & & & PELM\_A1 & N & Y & Y & 1.00 & 11131 & 25.50 & 13.89 & 16.00 & 1.15 & 9.13 & 163.15 \\
 & & & PELM\_A2 & Y & N & Y & 1.00 & 11257 & 22.20 & 15.01 & 15.02 & 1.00 & 8.51 & 150.51 \\
 & & & PELM\_A3 & Y & Y & N & 1.00 & 9440 & 27.20 & 12.91 & 14.05 & 1.09 & 8.65 & 182.43 \\
\cmidrule{3-15}
 & & \multirow{4}{*}{Moderate} & PELM & Y & Y & Y & 1.00 & 11098 & 25.20 & 13.82 & 15.70 & 1.14 & 8.92 & 159.89 \\
 & & & PELM\_A1 & N & Y & Y & 1.00 & 10993 & 25.20 & 13.46 & 16.00 & 1.19 & 9.13 & 165.25 \\
 & & & PELM\_A2 & Y & N & Y & 1.00 & 11162 & 23.80 & 14.30 & 14.30 & 1.00 & 8.75 & 155.99 \\
 & & & PELM\_A3 & Y & Y & N & 1.00 & 9227 & 26.70 & 12.93 & 15.02 & 1.16 & 8.22 & 177.23 \\
\cmidrule{3-15}
 & & \multirow{4}{*}{Low} & PELM & Y & Y & Y & 1.00 & 11130 & 25.90 & 13.03 & 15.48 & 1.19 & 9.13 & 163.27 \\
 & & & PELM\_A1 & N & Y & Y & 1.00 & 10850 & 25.20 & 13.67 & 16.00 & 1.18 & 9.16 & 168.05 \\
 & & & PELM\_A2 & Y & N & Y & 1.00 & 11772 & 22.90 & 14.81 & 14.84 & 1.00 & 9.35 & 158.13 \\
 & & & PELM\_A3 & Y & Y & N & 1.00 & 9995 & 26.20 & 13.60 & 15.68 & 1.15 & 8.88 & 176.74 \\
\cmidrule{3-15}
 & & \multirow{4}{*}{Minimum} & PELM & Y & Y & Y & 1.00 & 10807 & 25.50 & 13.56 & 15.72 & 1.16 & 8.86 & 163.12 \\
 & & & PELM\_A1 & N & Y & Y & 1.00 & 11209 & 23.60 & 14.21 & 16.00 & 1.12 & 9.16 & 162.63 \\
 & & & PELM\_A2 & Y & N & Y & 1.00 & 11558 & 23.00 & 14.89 & 14.91 & 1.00 & 8.27 & 142.34 \\
 & & & PELM\_A3 & Y & Y & N & 1.00 & 9761 & 25.90 & 13.69 & 15.18 & 1.11 & 8.71 & 177.67 \\
\midrule
\multirow{16}{*}{\rotatebox{90}{\textbf{Jetson Orin Nano}}} & \multirow{16}{*}{\textbf{1B}} & \multirow{4}{*}{High} & PELM & Y & Y & Y & 1.00 & 5688 & 17.60 & 13.13 & 14.42 & 0.91 & 8.51 & 297.56 \\
 & & & PELM\_A1 & N & Y & Y & 1.00 & 5926 & 16.30 & 14.64 & 16.00 & 1.09 & 9.18 & 308.42 \\
 & & & PELM\_A2 & Y & N & Y & 1.00 & 6005 & 16.60 & 14.85 & 14.83 & 1.00 & 9.64 & 319.50 \\
 & & & PELM\_A3 & Y & Y & N & 1.00 & 5652 & 19.50 & 14.30 & 15.64 & 1.10 & 8.97 & 315.65 \\
\cmidrule{3-15}
 & & \multirow{4}{*}{Moderate} & PELM & Y & Y & Y & 1.00 & 5486 & 19.60 & 11.96 & 13.05 & 0.92 & 7.80 & 283.03 \\
 & & & PELM\_A1 & N & Y & Y & 1.00 & 5988 & 16.90 & 14.54 & 16.00 & 1.10 & 9.14 & 303.78 \\
 & & & PELM\_A2 & Y & N & Y & 1.00 & 6009 & 17.00 & 14.32 & 14.33 & 1.00 & 9.00 & 298.15 \\
 & & & PELM\_A3 & Y & Y & N & 1.00 & 5263 & 20.70 & 12.89 & 13.53 & 1.05 & 8.85 & 334.64 \\
\cmidrule{3-15}
 & & \multirow{4}{*}{Low} & PELM & Y & Y & Y & 1.00 & 5218 & 19.80 & 11.98 & 12.69 & 0.94 & 8.28 & 315.67 \\
 & & & PELM\_A1 & N & Y & Y & 1.00 & 6010 & 16.70 & 14.11 & 16.00 & 1.14 & 9.13 & 302.26 \\
 & & & PELM\_A2 & Y & N & Y & 1.00 & 5362 & 17.90 & 13.69 & 13.69 & 1.00 & 9.70 & 360.09 \\
 & & & PELM\_A3 & Y & Y & N & 1.00 & 5253 & 20.70 & 12.80 & 13.67 & 1.06 & 9.14 & 346.40 \\
\cmidrule{3-15}
 & & \multirow{4}{*}{Minimum} & PELM & Y & Y & Y & 1.00 & 5486 & 17.10 & 11.96 & 13.19 & 0.91 & 8.34 & 302.58 \\
 & & & PELM\_A1 & N & Y & Y & 1.00 & 5906 & 14.20 & 13.92 & 16.00 & 1.15 & 9.07 & 305.77 \\
 & & & PELM\_A2 & Y & N & Y & 1.00 & 5209 & 16.60 & 12.94 & 12.94 & 1.00 & 7.52 & 287.44 \\
 & & & PELM\_A3 & Y & Y & N & 0.99 & 4937 & 21.20 & 12.19 & 13.01 & 1.06 & 7.44 & 298.47 \\
\bottomrule
\end{tabular}
}
\end{table}

We create three reduced versions of \sys for ablation studies to evaluate the efficacy of its individual components. \textbf{A1}: In this version, we remove the variable depth control from \sys, fixing the depth to its maximum value. \textbf{A2}: Here, we eliminate the self-speculative decoding component, such that \sys only controls processor frequency and the exit layer (via variable depth) during autoregressive decoding. \textbf{A3}: In this case, we remove frequency control from \sys and allow the BLTN governor to manage it, leaving \sys responsible solely for self-speculative decoding with variable depth. 

\rev{Table~\ref{tab:ablation_study_components} provides an explicit component-wise breakdown of \sys.
The three columns (DVFS, VD, SSD) correspond to three orthogonal control dimensions.
The full \sys enables all three (Y/Y/Y), while A1, A2, and A3 each disable exactly one component, forming a complete leave-one-out ablation matrix.
Results are grouped by thermal regimes to reveal component sensitivity under distinct hardware constraints.}

Removing Variable Depth (\textbf{A1}) universally degrades performance-per-joule (\texttt{PPJ}) in most scenarios. By being forced to use the maximum model depth, \textbf{A1} consistently performs more work per token (higher LPATm) and is universally slower than the full \sys~system. This indicates that dynamically selecting an earlier exit layer is critical for computational efficiency. The removal of Speculative Decoding (\textbf{A2}) causes the most significant performance degradation. This purely autoregressive version is simultaneously slower and consumes substantially more energy in every tested scenario (e.g., 8B Moderate: 31680\,J vs. \sys's 28059\,J), resulting in a significant drop in \texttt{PPJ}. This demonstrates that speculative decoding is essential to \sys's speed and energy savings. Finally, removing Frequency Control (\textbf{A3}) reveals a critical trade-off. For the light 1B workload on AGX Orin, \textbf{A3}  achieves superior \texttt{PPJ} by aggressively prioritizing speed. However, this naive strategy fails on all \textit{moderate-to-heavy} workloads. In those cases, \textbf{A3} consumes disproportionate energy for its speed gains (e.g., 35705\,J vs. 28059\,J for 8B Moderate), severely hurting efficiency. In conclusion, this study highlights the importance of \sys's components.

\subsection{Query-Locality Impact}

\begin{table}[h]
\centering
\caption{\rev{Metrics across different localities. (TSA=Task Score Average, LPATm=layers per accepted token mean value, SR=Speculation Ratio, PPJ=Performance per joule)}}
\label{tab:locality}
\resizebox{\columnwidth}{!}{
\begin{tabular}{ccccccccccc}
\toprule
\textbf{H (\#P)} & \textbf{Cooling} & \textbf{Locality} & \textbf{Energy} & \textbf{Speed} & \textbf{LPATm} & \textbf{VDm} & \textbf{SR} & \textbf{TSA} & \textbf{PPJ} \\
\midrule
\multirow{16}{*}{\rotatebox{90}{\textbf{AGX Orin (8B)}}} & \multirow{4}{*}{High} & 1 & 30934 & 9.2 & 26.07 & 29.62 & 1.14 & 20.71 & 133.26 \\
& & 10 & 27867 & 9.6 & 25.55 & 27.86 & 1.09 & 20.40 & 145.68 \\
& & 20 & 27547 & 9.5 & 24.93 & 29.32 & 1.18 & 20.84 & 150.55 \\
& & 40 & 30868 & 10.4 & 24.85 & 29.94 & 1.20 & 21.19 & 136.60 \\
\cmidrule{2-10}
& \multirow{4}{*}{Moderate} & 1 & 28059 & 9.6 & 24.84 & 28.55 & 1.15 & 22.69 & 160.91 \\
& & 10 & 28562 & 10.0 & 24.63 & 29.22 & 1.19 & 23.47 & 163.55 \\
& & 20 & 28753 & 9.8 & 26.14 & 30.15 & 1.15 & 20.51 & 141.95 \\
& & 40 & 29470 & 9.9 & 25.63 & 30.31 & 1.18 & 22.01 & 148.66 \\
\cmidrule{2-10}
& \multirow{4}{*}{Low} & 1 & 28360 & 9.2 & 23.95 & 26.93 & 1.12 & 21.36 & 149.90 \\
& & 10 & 26551 & 10.0 & 23.28 & 26.68 & 1.15 & 20.30 & 152.16 \\
& & 20 & 27626 & 9.6 & 23.76 & 27.43 & 1.15 & 22.48 & 161.93 \\
& & 40 & 26357 & 10.6 & 23.22 & 27.61 & 1.19 & 19.05 & 143.86 \\
\cmidrule{2-10}
& \multirow{4}{*}{Minimum} & 1 & 29617 & 9.3 & 23.32 & 26.86 & 1.15 & 19.20 & 128.99 \\
& & 10 & 27272 & 9.9 & 23.07 & 26.02 & 1.12 & 19.13 & 139.55 \\
& & 20 & 27711 & 9.9 & 23.61 & 26.90 & 1.14 & 19.84 & 142.47 \\
& & 40 & 27258 & 10.7 & 23.00 & 26.68 & 1.16 & 18.83 & 137.45 \\
\midrule
\multirow{16}{*}{\rotatebox{90}{\textbf{Orin Nano (1B)}}} & \multirow{4}{*}{High} & 1 & 5688 & 17.6 & 13.13 & 14.42 & 0.91 & 8.51 & 297.56 \\
& & 10 & 5681 & 17.4 & 13.31 & 14.86 & 1.11 & 9.88 & 346.14 \\
& & 20 & 5514 & 17.7 & 12.74 & 14.25 & 1.12 & 8.68 & 313.36 \\
& & 40 & 5562 & 18.3 & 12.96 & 14.18 & 1.10 & 8.48 & 303.29 \\
\cmidrule{2-10}
& \multirow{4}{*}{Moderate} & 1 & 5486 & 19.6 & 11.96 & 13.05 & 0.92 & 7.80 & 283.03 \\
& & 10 & 5729 & 17.6 & 13.36 & 15.06 & 1.12 & 9.08 & 315.24 \\
& & 20 & 5251 & 18.4 & 12.37 & 13.42 & 1.09 & 8.46 & 320.74 \\
& & 40 & 5321 & 17.9 & 12.70 & 14.25 & 1.12 & 9.42 & 352.36 \\
\cmidrule{2-10}
& \multirow{4}{*}{Low} & 1 & 5218 & 19.8 & 11.98 & 12.69 & 0.94 & 8.28 & 315.67 \\
& & 10 & 5445 & 18.3 & 12.50 & 13.80 & 1.10 & 8.70 & 318.02 \\
& & 20 & 5267 & 18.3 & 11.86 & 13.35 & 1.12 & 8.83 & 333.44 \\
& & 40 & 5096 & 19.6 & 11.38 & 12.85 & 1.12 & 7.59 & 296.55 \\
\cmidrule{2-10}
& \multirow{4}{*}{Minimum} & 1 & 5486 & 17.1 & 11.96 & 13.19 & 0.91 & 8.34 & 302.58 \\
& & 10 & 5487 & 16.9 & 11.69 & 13.09 & 1.12 & 7.65 & 277.42 \\
& & 20 & 5309 & 18.4 & 11.30 & 12.68 & 1.12 & 8.13 & 304.81 \\
& & 40 & 4925 & 18.6 & 11.14 & 12.80 & 1.15 & 8.15 & 329.46 \\
\bottomrule
\end{tabular}
}
\end{table}

While all previous evaluations used a randomized query dataset, real-world scenarios often exhibit temporal locality, with consecutive queries belonging to the same task type \cite{frieder2024caching}. To investigate the impact of locality on end-to-end performance, we reordered the query dataset to form clusters of queries from the same task domain and did evaluations on 8B@AGX Orin and 1B@Orin Nano.

The results are in shown in Table~\ref{tab:locality}. \sys intelligently adapts to and exploits query locality. Rather than a static system, it learns from the locality in query patterns for more effective speculations. The primary finding is that speculation speedup ratio (SR) likely benefits directly from clustered query sequence. For instance, 18 out of 24 cases for larger locality (10, 20, 40) show an increase in SR compared to fully randomized order (Locality 1), which contributes to lower LPATm and improved end-to-end decoding speed. As a result, the total energy consumption is reduced for larger locality for most cases.

However, the results also reveal complexities, particularly for 1B model on the Nano. In that setting, 8 out of 12 higher-locality cases do not result in higher speedups, in contrast to 8B model. Moreover, task performance (TSA) exhibits greater variance than 8B on Jetson. We hypothesize this is due to the constraints of the small (1B) model, which makes correct speculation more challenging. While SR may increase on plausible tokens, the final task quality could be impacted. This, in turn, negatively affects the PPJ score. Despite these complex and non-monotonic behaviors, \sys does not exhibit catastrophic performance degradation. This overall stability underscores its robustness and usability across diverse query locality patterns.

\subsection{\rev{System Overhead}}

\rev{We empirically measure the runtime, power, and memory overhead of PELM, and show that its control and caching mechanisms introduce negligible system cost.}

\rev{PELM’s DQN agent is lightweight and CPU-only. The policy network (BranchQNet) is a small MLP with two 32-wide fully connected layers and shallow linear heads (5{,}687 parameters, $\approx$22\,KB in FP32), and all action selection and training run on CPU without launching GPU kernels. At a 10\,Hz control rate (100\,ms period), action selection takes 0.087--0.922\,ms and training updates 9--18\,ms, remaining well within the control window. End-to-end measurements show that PELM does not increase CPU power relative to baselines ($\approx$1.8--2.2\,W), while substantially reducing GPU power (e.g., 12.39\,W vs.\ 20--21\,W on 8B@AGX, Moderate), indicating that DRL overhead is negligible compared to the energy savings it enables.}

\rev{HQCache stores hidden states (not KV cache) and incurs small latency and memory cost. On AGX Orin, HQCache resume/load latency is 0.090--0.320\,ms for token lengths 1--1024, whereas a single decoder-layer forward at the same length takes 4.195--20.416\,ms ($\approx$46--60$\times$ slower), making cache loading far cheaper than recomputation. For an 8B LLaMA-style model with hidden size $d=4096$ in FP16, one checkpoint of length $L$ costs $L \times d \times 2$ bytes (16\,MiB at $L=2048$; 128\,MiB for $K=8$), typically $<1\%$ of FP16 model weights and modest relative to Jetson unified memory.}

%% file: sections/6-discussion.tex
\section{Discussion}
\label{sec-discussion}

\noindent\textbf{Limitations.} First, the effectiveness of \sys depends on access to LLMs that expose intermediate-layer outputs through early-exit supervision (e.g., LayerSkip~\cite{elhoushi2024layerskip}). Currently, publicly available models with robust early-exit heads are concentrated within the LLaMA family. \rev{Extending \sys to other model families requires enabling intermediate layer supervision during pretraining or fine-tuning to expose early-exit logits. This modification is confined to the model training pipeline through intermediate layer supervision (SFT); the system-level components of \sys --— including the DVFS interface, runtime telemetry, and closed-loop governor —-- remain unchanged. When early-exit capability is unavailable, \sys degrades to a reduced configuration that combines DVFS control with dynamic self-speculative decoding using full-depth verification (see ablation study in Section~\ref{sec-results-ablation}). In this case, verification depth control is disabled, while remaining speculation control, hardware-level adaptation, and QoE-aware regulation continue to operate.}

\rev{Quantization is widely adopted for on-device LLM deployment and represents a complementary optimization to our approach. In this work, we focus on unquantized checkpoints to isolate and analyze the cross-layer control behavior of \sys. Integrating quantized models is feasible but requires additional model preparation and validation, particularly to ensure consistent intermediate-layer outputs under variable-depth execution and HQCache state resumption. From a control perspective, quantization primarily reshapes the latency–power characteristics of the model, which the closed-loop governor is designed to adapt to in principle; however, a rigorous quantized evaluation remains future work. We view this as a promising direction to further expand the applicability of \sys to larger-scale models within edge device constraints.}

Our evaluation is also confined to NVIDIA Jetson platforms and their DVFS interfaces; portability to other SoCs may require remapping control knobs (e.g., heterogeneous CPU/GPU/NPU governors) and re-tuning policy targets.

\noindent\textbf{Future Directions.} \add{Promising future directions include making the governor more input- and output-aware. Token-level difficulty signals, such as entropy or probability margin, could enable finer-grained depth control, while lightweight task classification could guide verification depth selection based on task complexity, improving the balance between hardware constraints and task-level quality under tight thermal and power budgets. Another direction is to improve decoding algorithms for on-device LLMs through hardware-software co-design. This includes more effective training of early-exit LLMs to improve exit reliability and reduce verification cost, extending such schemes to other model families such as Phi~\cite{abdin2024phi} and Qwen~\cite{yang2025qwen3}, and developing new methods to accelerate on-device generation.}

\rev{An important future direction is to evaluate \sys under co-running background workloads and platform-wide DVFS policies. In practical deployments, system-level governors and concurrent services may introduce additional contention and dynamic frequency scaling interactions; studying joint control or coordination mechanisms would further strengthen robustness in multi-tenant edge environments.}

\add{Lastly, many orthogonal yet useful techniques such as quantization, and memory offloading could be incorporated into \sys for deploying LLM on broader platforms such as Raspberry Pi-class SBCs, smartphones, and NPUs, which will also require adapting to heterogeneous DVFS controls and thermal envelopes.}

%% file: sections/8-conclusion.tex
\section{Conclusion}

This paper proposes \sys, a novel power-efficient optimization framework co-designing DVFS and speculative decoding for on-device LLM inference. \sys~bridges the gap between prior decoding-agnostic DVFS methods and hardware-agnostic speculative decoding. By introducing variable depth for verification execution, it enlarges the optimization space for further efficiency gains. We evaluate it with diverse datasets across different devices, hardware constraints, and QoE requirements, verifying its efficiency.